%% file: egpaper_for_review.tex
\documentclass[10pt,twocolumn,letterpaper]{article}

\usepackage{iccv}
\usepackage{times}
\usepackage{epsfig}
\usepackage{graphicx}
\usepackage{amsmath}
\usepackage{amssymb}

\usepackage{booktabs}
\usepackage{multirow}
\usepackage{graphics}
\usepackage{float}
\usepackage{wrapfig}
\usepackage{url}
\usepackage{subcaption}
\usepackage{soul}
\usepackage{comment}
\usepackage[table,xcdraw, dvipsnames]{xcolor}

\usepackage[pagebackref=true,breaklinks=true,letterpaper=true,colorlinks,bookmarks=false]{hyperref}

\iccvfinalcopy %

\def\iccvPaperID{8615} %

\ificcvfinal\fi

\newcommand\blfootnote[1]{%
  \begingroup
  \renewcommand\thefootnote{}\footnote{#1}%
  \addtocounter{footnote}{-1}%
  \endgroup
}

\begin{document}

\title{{\color{Salmon}{MINOTAUR}}: {\color{Salmon}{M}}ulti-task V{\color{Salmon}{i}}deo Grou{\color{Salmon}{n}}ding Fr{\color{Salmon}{o}}m Mul{\color{Salmon}{t}}imod{\color{Salmon}{a}}l Q{\color{Salmon}{u}}e{\color{Salmon}{r}}ies}

\author{
Raghav Goyal$^{*1,2}$ \hspace{0.25in} Effrosyni Mavroudi$^{4}$ \hspace{0.25in} Xitong Yang$^{4}$ \hspace{0.25in} Sainbayar Sukhbaatar$^{4}$ \\ Leonid Sigal$^{1,2,3}$ \hspace{0.25in} Matt Feiszli$^{4}$ \hspace{0.25in} Lorenzo Torresani$^{4}$ \hspace{0.25in} Du Tran$^{4}$\\ \\
$^1$University of British Columbia \hspace{0.3in}
$^2$Vector Institute for AI \hspace{0.5in}   $^3$CIFAR AI Chair \\
$^4$ Meta AI
}
\vspace{-4pt}

\ificcvfinal\thispagestyle{empty}\fi

\maketitle

\begin{abstract}
Video understanding tasks take many forms, from action detection to visual query localization and spatio-temporal grounding of sentences. These tasks differ in the type of inputs (only video, or video-query pair where query is an image region or sentence)
and outputs (temporal segments or spatio-temporal tubes). However, at their core they require the same fundamental understanding of the video, i.e., the actors and objects in it, their actions and interactions.
So far these tasks have been tackled in isolation with individual, highly specialized architectures and pre-extracted features, which do not exploit the interplay between tasks. In contrast, in this paper, we present a single, unified model for tackling query-based video understanding in long-form videos. In particular, our model can address all three tasks of the Ego4D Episodic Memory benchmark which entail queries of three different forms: given an egocentric video and a {\em visual}, {\em textual} or {\em activity} query, the goal is to determine when and where the answer can be seen within the video. Our model design is inspired by recent query-based approaches to spatio-temporal grounding, and contains modality-specific query encoders and task-specific sliding window inference
that allow multi-task training with diverse input modalities and different structured outputs.
We exhaustively analyze relationships
among the tasks and illustrate that 
cross-task learning leads to improved performance on each individual task, as well as the ability to generalize to unseen tasks, such as zero-shot spatial localization of language queries.

\blfootnote{* Algorithmic work was done during an internship at Meta AI.}
\blfootnote{$\ \ $Application of the model to some down-stream tasks was conducted}
\blfootnote{$\ \ $at UBC / Vector in collaboration with Meta AI partners.}
\blfootnote{$\ \ $Correspondence: \texttt{rgoyal14@cs.ubc.ca}}

\end{abstract}

\input{introduction}

\input{related_work}

\input{approach}

\input{experiments}

\input{conclusion}

\section*{Acknowledgement}
{\small This work was initiated as an internship project at Meta AI and has been to a large part enabled and made possible by the computational resources provided by Meta. In addition, this work was funded, in part, by the Vector Institute for AI, Canada CIFAR AI Chair, NSERC CRC and an NSERC DG. Hardware resources, in part, were provided by the Province of Ontario, the Government of Canada through CIFAR, and companies sponsoring the Vector Institute\footnote{\url{www.vectorinstitute.ai/\#partners}}. Finally, we would like to thank Mennatullah Siam for valuable discussions.}

{\small
\bibliographystyle{ieee_fullname}
\bibliography{egbib}
}

\input{appendix}

\end{document}

%% file: introduction.tex
\begin{figure}[t]
  \centering
   \includegraphics[width=0.95\linewidth]{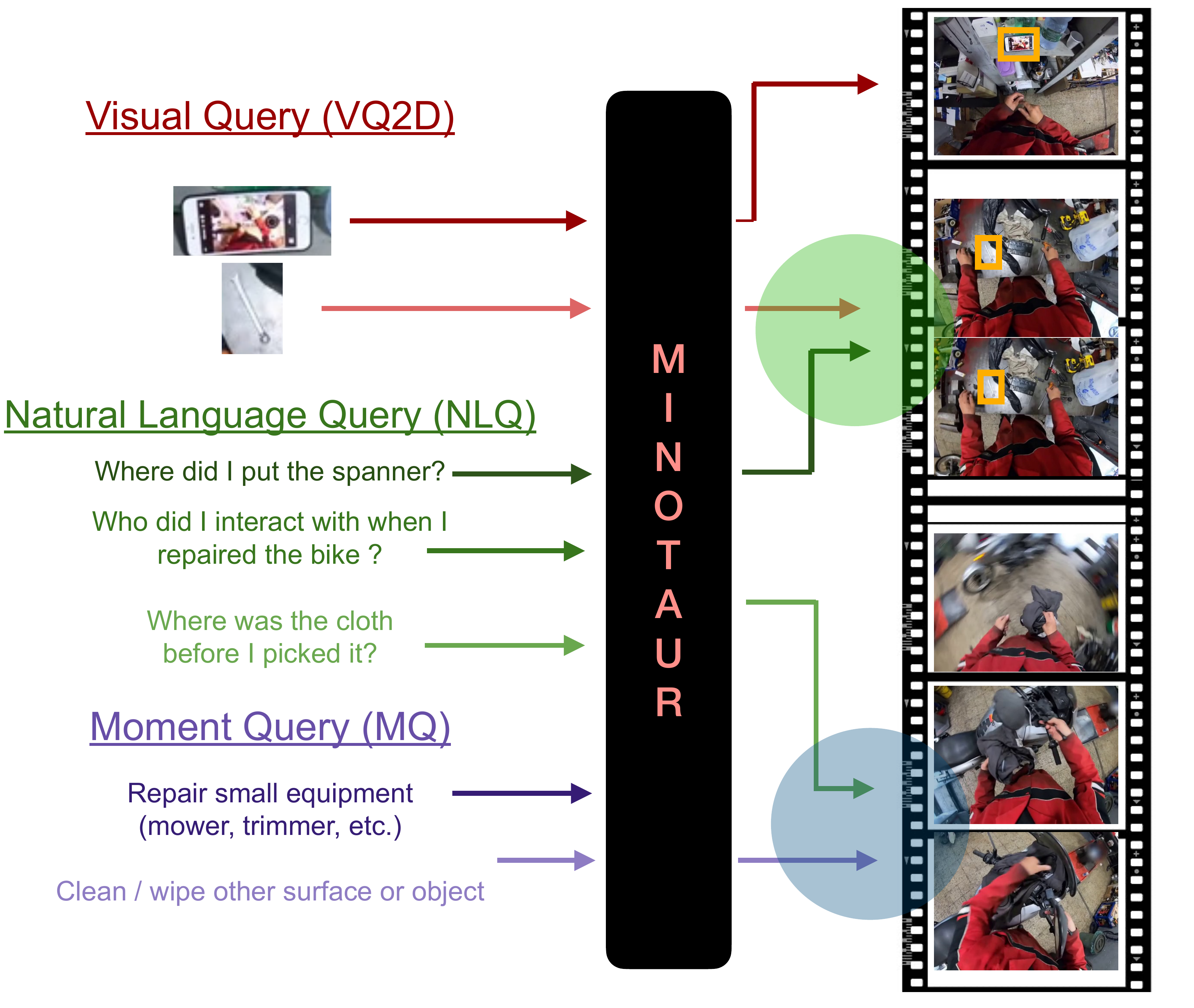}
    \vspace{-6pt}
    
   \caption{\textbf{Overview.} Our unified approach tackles the three episodic memory tasks with a single model architecture, MINOTAUR, and leverage the complementary information across different tasks (depicted with shaded circles).}
   
   \label{fig:intro_fig}
   \vspace{-16pt}
\end{figure}

\section{Introduction}
\label{sec:intro}

Fine-grained video understanding has emerged as a crucial capability for the development of augmented reality (AR) and robotics applications. To achieve this level of video comprehension, an agent (\eg, virtual assistant) must be equipped with the ability to identify and reason about events and objects captured in videos, tackling a range of vision tasks such as activity detection~\cite{caba2015activitynet,sigurdsson2016hollywood,zhao2017temporal}, object retrieval~\cite{grauman2022ego4d},%
and (spatio-)temporal grounding~\cite{anne2017localizing,chen2018temporally,gao2017tall,lei2021detecting}. 
However, current research on video understanding has focused primarily on training individual, highly specialized models using task-specific annotated datasets~\cite{caba2015activitynet,gao2017tall,gu2018ava,kay2017kinetics,regneri2013grounding,zhang2020does}, without considering the synergies between multiple tasks and the model's generalization to novel tasks.

In this paper, we seek an alternative approach -- building a \emph{unified} framework for multiple fine-grained video understanding tasks. Specifically, we aim to investigate two key research questions: (i) Is it feasible to design a single architecture that can seamlessly handle multiple tasks with multimodal inputs in an end-to-end manner? (ii) To what extent can this unified framework leverage the complementary information present between different tasks to achieve superior performance and generalization?

Intuitively, the learning process of a model can be facilitated by capturing the underlying relationships between different tasks. For example, a model trained to answer natural language queries~\cite{grauman2022ego4d} (\eg, ``{\em Where did I put the spanner?}") can benefit from the learning of answering visual queries (\eg, searching the latest occurrence of ``{\em spanner}") and moment queries (\eg, ``{\em repair small equipment (mower, trimmer, etc.)}"), as illustrated in Figure~\ref{fig:intro_fig}. We show similar empirical findings with our unified framework in Section~\ref{sec:results}. Moreover, a model trained across multiple tasks has been observed to exhibit superior generalization ability~\cite{lu202012,lu2022unified} (including to novel tasks as we show in Section~\ref{sec:results}) due to the fact that the model has access to diverse data annotated for different purposes and is not overfitted to a single objective. 
Our goals in this paper are synergetic with those of video foundation models \cite{chen2022internvideo,lin2022egocentric}, but are also distinct in that our focus is not on learning a model (using pretex tasks) that can easily adapt to various target tasks, but rather one {\em joint} model that is applicable and can solve variety of target video understanding tasks directly.

While the idea of building a unified framework for multiple tasks is conceptually straightforward, it does come with a number of core technical challenges. The first challenge is the inconsistent input (visual, lingual)/output (spatial, temporal) format and output distribution across different tasks. Taking the three episodic memory tasks~\cite{grauman2022ego4d} as an example, the input of natural language query (NLQ) is free-form language and the output spans $< 2$ seconds, while the input of visual query (VQ) is an image crop and the output of moment query (MQ) typically spans $>1$ minute. The problem becomes more severe when localizing a short answer in the long-form videos (8 minutes) in an end-to-end manner. 
In addition, effective training of models that capture multimodal information across tasks without compromising per-task performance remains an open challenge~\cite{Malhotra2022, wang2019}.

In light of this, we present {\color{Salmon}MINOTAUR}, a unified multi-task model that is designed to accept long videos and multi-modal queries as input, and generate diverse structured outputs. The key idea is to cast diverse tasks as special cases of {\em (spatio-)temporal grounding of queries}, which allows them to be streamlined towards predicting the spatial and/or temporal response locations of the queries within videos.
MINOTAUR builds upon the Transformer encoder-decoder architecture~\cite{lu202012,yang2022tubedetr,jaegle2021perceiver}, and task-specificity is baked into the input/output processing modules such as modality-specific query embeddings and prediction heads.
To address the challenges arising from inconsistent output distribution in long videos, we design a sliding-window based training and inference regime, combined with a foreground frame prediction module capable of generalizing across different tasks.
Unlike prior work on long video understanding that relies on pre-extracted visual features~\cite{zhang2022actionformer,zhao2021video,lin2022egocentric}, our model is fully end-to-end learnable and can be trained with partially annotated data, {\em i.e.}, without requiring annotations for all tasks per video.
We empirically illustrate that our joint model is capable of transferring and utilizing information across tasks (see Table~\ref{tab:multi-task-val-short}) improving performance by as much as 18\% on NLQ task compared to the single-task counterpart. We also show that our model can generalize and produce meaningful results for zero-shot spatio-temporal grounding, without training on the task.

\vspace{0.05in}
\noindent
{\bf Contributions.}
The contributions and novelty of this work is both methodological and technical. 
Foremost, we propose MINOTAUR, a unified, Transformer-based model for grounding multi-modal queries in long-form videos. To the best of our knowledge, this is the first work that trains and evaluates a single model for tasks ranging from spatio-temporal grounding to activity detection. We illustrate that this architecture is able to learn and leverage information across varied tasks as well as enable novel zero-shot task in inference. 
In service of this model, we propose efficient, multi-task training and inference strategies, which enable training with fixed duration, partially annotated video segments, and testing on variable-length, long-form videos. 
We train our model in both single-task and multi-task settings and evaluate it on the three tasks of the Ego4D Episodic Memory benchmark~\cite{grauman2022ego4d}. Our results show that our unified model is competitive with architectures designed and optimized for each individual task, as well as allows zero-shot spatio-temporal grounding (not shown elsewhere).

%% file: related_work.tex
\section{Related Work}
\label{sec:related-work}
Our work is related to prior works on fine-grained video understanding (\eg, action detection, video grounding), multi-task learning and video-language modeling.

\vspace{0.05in}
\noindent
{\bf Video grounding.} Grounding natural language query in a reference video has been an active research area in recent years. The task can be categorized to temporal grounding~\cite{chen2018temporally,chen2019localizing,he2019read,lin2020weakly,mithun2019weakly,rodriguez2020proposal,wang2020temporally,wang2019language,zhang2020learning,soldan2021vlg,hou2022cone} which aims to localizing temporal segments in a video, and spatio-temporal grounding~\cite{jain2014action,pan2021actor,tang2020asynchronous,tran2012max,weinzaepfel2015learning} which further requires localizing spatio-temporal tubes.

One direction for temporal grounding is the proposal-based approach \cite{anne2017localizing, gao2017tall, chen2019semantic, xu2019multilevel, zhang2020span}, where proposals are first generated and then ranked according to query.
The proposal-free approach\cite{ghosh2019excl,yuan2019find,zhang2020span,lei2021detecting}, on the other hand,  predicts the start and end positions directly given the query input.
For spatio-temporal grounding, prior works approach the task from a frame-level perspective \cite{peng2016multi,saha2016deep,singh2017online,weinzaepfel2015learning,chen2021watch,pan2021actor} where predictions are made on individual frames and linked to form action tubes, or tubelet-level perspective \cite{jain2014action,hou2017tube,li2018recurrent,yang2019step,zhao2019dance,li2020actions} where they directly treat tubelet as a unit. Recent works \cite{yang2022tubedetr,zhao2022tuber}, highlight the effectiveness of Transformer-based approaches, where in particular,
TubeDETR \cite{yang2022tubedetr} proposed a spatio-temporal grounding system for a given language query. We also use a Transformer-based approach, extending TubeDETR \cite{yang2022tubedetr} to include task-specific components to enable multi-task learning.

\vspace{0.05in}
\noindent
{\bf Temporal action detection.} The task aims to determine the semantic label and the temporal interval of every action instance in an untrimmed video \cite{caba2015activitynet,sigurdsson2016hollywood,lin2018bsn,xu2017r,zhao2021video}. Similar to temporal grounding, prior works either adopt a two-stage approach which proposes and classifies action proposals using graph neural networks \cite{bai2020boundary,xu2020g,zeng2019graph,zhao2021video} or Transformers \cite{tan2021relaxed,sridhar2021class}, or a single-stage approach that localizes actions in one step \cite{lin2017single,lin2021learning,yang2020revisiting}. Recent single-stage models leverage the Transformer-based framework and achieves the state-of-the-art results~\cite{zhang2022actionformer,liu2022end}. Our work also follows a single-stage approach where a dedicated \textit{foreground} head is introduced to guide the detection in long videos.

It is also noteworthy that most temporal localization or grounding methods rely on pre-extracted visual features to model long videos and, in doing so, are limited in  
ability to holistically understand videos. In this work, we propose an end-to-end learning framework that tackles all the above tasks in a unified manner.

\vspace{0.05in}
\noindent
{\bf Multi-task learning.} Multi-task learning has been applied to vision \cite{bragman2019stochastic,misra2016cross,strezoski2019many,zhang2013robust}, language \cite{raffel2020exploring,collobert2008unified,liu2019multi,mccann2018natural} and robotics \cite{parisotto2015actor,teh2017distral}. Of particular interest are works that explore multi-modal multi-task learning 
\cite{nguyen2019multi,pramanik2019omninet,lu202012}. In particular, Lu \textit{et al.} \cite{lu202012} trains on $12$ image-understanding tasks with task-specific prediction heads. In our work, we also use task-specific components for encoding query and decoding outputs, but focus on video understanding. 

\vspace{0.05in}
\noindent
{\bf Vision and Language.} Recently, vision-language approaches \cite{lu2019vilbert,tan2019lxmert,su2019vl,li2020oscar,kamath2021mdetr,radford2021learning,li2022blip} using contrastive or masked-data-modeling pre-training losses have proven to be effective for downstream tasks, which are further extended to videos \cite{li2022align}, and in particular, EgoVLP \cite{lin2022egocentric} and InternVideo \cite{chen2022internvideo} extends the video-language pre-training to egocentric domain. However, our work specifically investigates a general spatio-temporal approach for videos. Another line of work looks at \textit{unified} models for structured tasks in vision \cite{kamath2021mdetr,gupta2022towards,alayrac2022flamingo,jaegle2021perceiver,lu2022unified,zhu2022uni}. Our approach follows the same line of work and proposes a unified framework for general spatio-temporal video understanding.

%% file: approach.tex
\vspace{-4pt}
\section{Approach}
\label{sec:approach}
\begin{figure*}[t]
  \centering
   \includegraphics[width=0.8\linewidth]{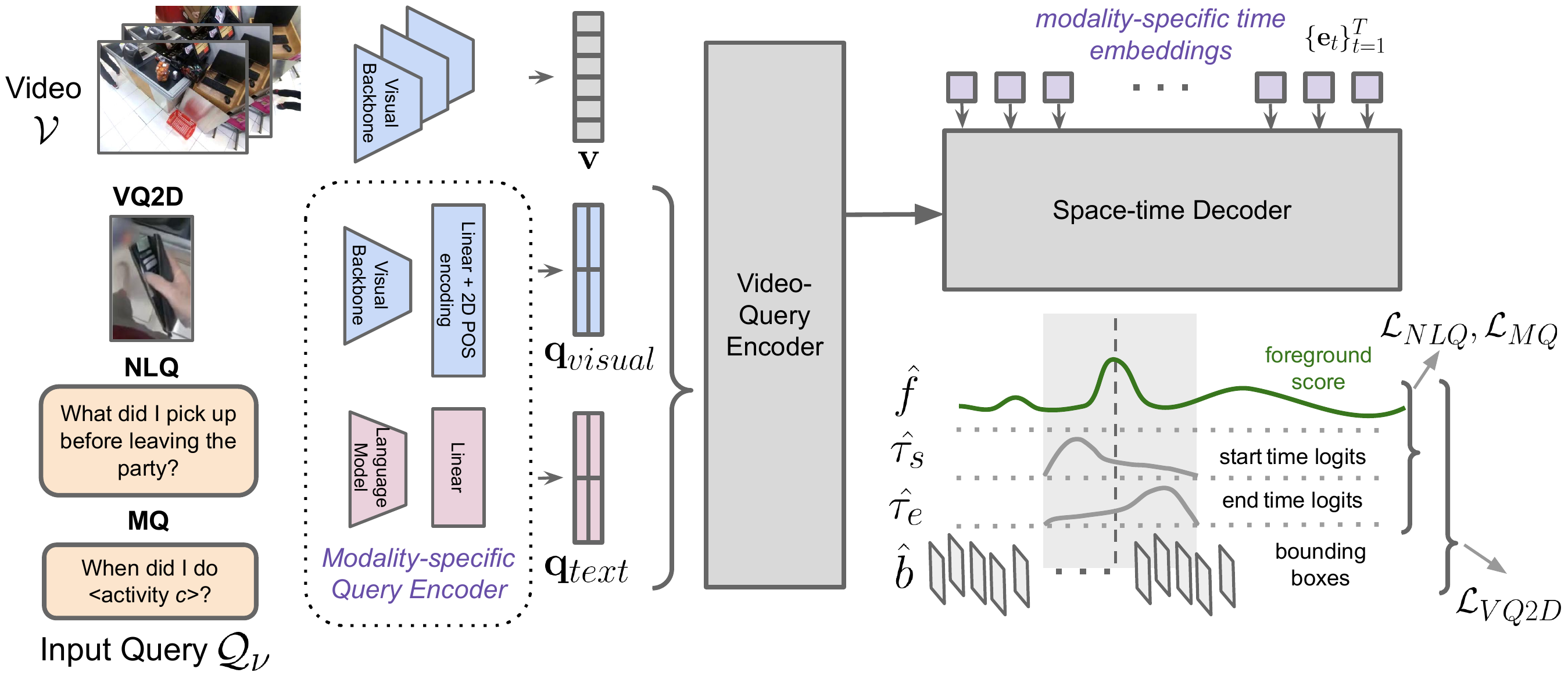}
    \vspace{-4pt}
   \caption{\textbf{Overview of the model} Given a video $\mathcal{V}$ and task-query $\mathcal{Q}_\nu$, our model encodes them using a Visual Backbone and Modality-specific Query Encoder. The obtained video and query features, $\mathbf{v}$ and $\mathbf{q}$, are encoded by Video-Query Encoder and decoded using Space-time Decoder by refining Modality-specific time embeddings $\{ \mathbf{e}_t \}_{t=1}^T$. Finally, the prediction heads predict foreground, start/end time logit, and bounding box for every frame. }
   \label{fig:overall-main}
   \vspace{-4pt}
\end{figure*}

\vspace{-4pt}
Our goal is to design a \emph{unified} framework that addresses various fine-grained video understanding tasks and leverages the complementary information they provide. Specifically, we focus on tackling the challenging episodic memory tasks~\cite{grauman2022ego4d} in this paper, which involves diverse types of queries that ask about a user's past experience.

\subsection{Problem Formulation}
The episodic memory tasks~\cite{grauman2022ego4d} include three different types of queries: (1) Visual Query (VQ2D) which asks for the most recent spatio-temporal location of the query object in a video; (2) Moment Query (MQ) which requires localizing all temporal segments of the query moment; (3) Natural Language Query (NLQ) which requires retrieving the temporal moments in a video that correspond to the input language question.
Due to their distinct input/output format and distribution, prior work typically approaches these tasks separately with specialized models (\eg, object-based siamese networks~\cite{grauman2022ego4d,xu2022my} for VQ2D, action detection models~\cite{zhang2022actionformer,zhao2021video} for MQ and temporal grounding networks~\cite{zhang2020span,lei2021detecting} for NLQ).
In contrast, our key idea is to unify all the three episodic memory tasks to a \emph{multi-task, multi-modal video grounding} problem.

Formally, given a reference video $\mathcal{V}$ and a query  $\mathcal{Q}_{\nu}$ where $\nu \in \{ \text{VQ2D}, \text{NLQ}, \text{MQ} \}$, our goal is to identify a response track $\mathcal{R}_{\nu}$ represented as a spatio-temporal tube or a temporal segment. Specifically,
\begin{itemize}
\setlength\itemsep{-0.12em}
    \item \textbf{VQ2D.} 
    With $\mathcal{Q}_{\nu}$ as a visual crop of an object at a query frame, %
    $\mathcal{R}_{\nu} = \{ r_s, r_{s+1}, ..., r_{e} \}$ is a spatio-temporal tube tracking the object of interest, where $s$ and $e$ are start and end frame indices respectively, and $r_i$ is a bounding box $(x, y, w, h) \in \mathbb{R}^4$ in the $i^{th}$ frame. If there are multiple occurrences of the object, the most recent occurrence before the query frame is retrieved.
    \item \textbf{MQ.} With $\mathcal{Q}_{\nu}$ being one of the activities from a pre-defined taxonomy (described in language), the response track consists of all the instances of query activity in the video, $\mathcal{R}_{\nu} = \{ (s_n, e_n) \}_{n=1}^N$, where $N$ is the total number of instances, $s_n$ and $e_n$ are start and end frame indices for the $n^{th}$ instance.
    \item \textbf{NLQ.} With $\mathcal{Q}_{\nu}$ as a text question, $\mathcal{R}_{\nu} = (s, e)$ is a temporal segment with $s$ and $e$ as start and end frame indices, respectively.

\end{itemize}

\subsection{MINOTAUR}

To jointly solve these three episodic memory tasks with a unified model, we introduce MINOTAUR, a Transformer-based architecture, illustrated in Figure~\ref{fig:overall-main}. Our model receives an egocentric video and a query (visual, textual or categorical) as inputs, and encodes them with a \emph{Visual Backbone} and a \emph{Modality-Specific Query Encoder}, respectively. After obtaining the video and query features, they are fused with a \emph{Video-Query Encoder} that captures multi-modal interactions. Then, the context-aware video-query features are decoded using a \emph{Space-Time Decoder}, which models long-range temporal interactions. The decoded features are fed to \emph{prediction heads} to generate a bounding box prediction, as well as predictions for the start, end, and foreground probabilities at each frame index. Lastly, these predictions are processed by \emph{task-specific output processing} modules, which convert them to %
response track $\mathcal{R}_{\nu}$ for each task.

We base our architecture on TubeDETR~\cite{yang2022tubedetr}, a state-of-the-art architecture for spatio-temporal grounding of natural language queries. Our model generalizes TubeDETR to handle multiple input query modalities, together with task-specific outputs in a long-form video, where the outputs can take the form of a single or multiple instances of spatio-temporal or temporal tubes. We describe %
TubeDETR \cite{yang2022tubedetr} below and our task-specific components in Section \ref{sec:multi-task-learn-infer}.

\vspace{0.07in}
\noindent
{\bf Visual backbone.} We start with encoding $T$ frames of video $\mathcal{V}$ using a 2D-CNN backbone and a feed-forward layer. We add 2D positional encoding to the features and flatten them to give $\mathbf{v} \in \mathbb{R}^{T\times HW \times d}$, where $T$ is the number of frames, $H$ and $W$ are height and width of feature map and $d$ is the hidden size of the model.

\vspace{0.06in}
\noindent
{\bf Video-Query encoder.}
Given input query features $\mathbf{q} \in \mathbb{R}^{L \times d}$ from our Modality-specific Query Encoder (described later in Section \ref{sec:multi-task-learn-infer}), where $d$ is the hidden size of the model and $L$ is the size of the query features, we replicate $\mathbf{q}$ across $T$ frames, concatenate with video features $\mathbf{v}$, and forward it through a $N_e$-layer transformer encoder to give $Enc(\mathbf{v}, \mathbf{q}) \in \mathbb{R}^{T \times (HW + L) \times d}$. The encoder applies transformer layers to each frame-query feature independently and effectively fuses query information with every frame feature. Note that for computational reasons, the video-query encoder can be applied to downsampled video-query features (sampled at regular intervals with stride $k$) and the output can be temporally replicated or upsampled by the factor of $k$ to match the original $T$ number of frames.

\vspace{0.07in}
\noindent
{\bf Space-time decoder.} The decoder takes as input a sequence of $T$ learned, modality-specific time embeddings $\{ \mathbf{e}_t \}_{t=1}^T \in \mathbb{R}^{d}$ (described later in Section \ref{sec:multi-task-learn-infer}). The decoder then refines the time embeddings for each frame $t$ by cross-attending to video-query encoder features $Enc(\mathbf{v}, \mathbf{q})$ in order to predict whether the answer to the query is visible at each frame $t$. Following TubeDETR \cite{yang2022tubedetr}, decoding is factorized over time and space for efficiency by having $N_d$-blocks
of temporal self-attention and frame-wise cross-attention layers, interleaved with feed-forward and normalization layers. The temporal self-attention layer allows time embeddings to attend to each other, thereby facilitating temporal interactions across the video. The frame-wise cross-attention layer 
performs cross-attention on each frame separately, where for an $t^{th}$ frame, the corresponding time embedding $\mathbf{e}_{t}$ cross-attends to its video-query encoder feature $Enc(\mathbf{v}, \mathbf{q})_i \in \mathbb{R}^{(HW + L) \times d}$. In effect, the decoder accounts for information in both temporal and spatial dimensions to produce refined time embeddings. See \cite{yang2022tubedetr} for more details. Finally, the refined time embeddings $\{ \hat{\mathbf{e}}_t \}_{t=1}^T$ are used for making predictions for every frame in the video.

\vspace{-0.1in}
\subsubsection{Modality-specific Encoders and Inference}
\label{sec:multi-task-learn-infer}
\vspace{-0.01in}
In this section, we describe the novel components of our architecture, inference procedures and losses that enable us to train and employ a single, unified model for all %
tasks.

\vspace{0.07in}
\noindent
{\bf Modality-specific query encoder.}
Depending on the task, the query $\mathcal{Q}_{\nu}$ can be a text question in the case of MQ and NLQ, or a visual crop in the form of image in the case of VQ2D. To handle different query modalities, we introduce a modality-specific query encoder. 
Language queries are embedded using a RoBERTa language model and mapped to hidden size $d$ of the model using a feed-forward layer, $\mathbf{q}_{text} \in \mathbb{R}^{L' \times d}$, where $L'$ is the number of tokens. On the other hand, %
visual queries are encoded using the visual backbone and a feed-forward layer. The resulting features are flattened along spatial dimensions to give $\mathbf{q}_{visual} \in \mathbb{R}^{H'W' \times d}$, where $H'$, $W'$ are spatial dimensions of the feature map. Formally, the query feature can be written as $\mathbf{q} \in \mathbb{R}^{L \times d}$, where $L \in \{L', H'W' \}$.

\vspace{0.07in}
\noindent
{\bf Modality-specific time embeddings.} As described earlier, the Space-time decoder takes as input a sequence of time embeddings $\{ \mathbf{e}_t \}_{t=1}^T$ and refines them using the encoded video-query features. We obtain the above modality-specific time embeddings by replicating a learned, modality-specific encoding vector ($\mathbf{e}^{text}$ or $\mathbf{e}^{visual}$) across time $T$ and summing it with a sinusoidal time-encoding.

\begin{figure}[t]
  \centering
   \includegraphics[width=\linewidth]{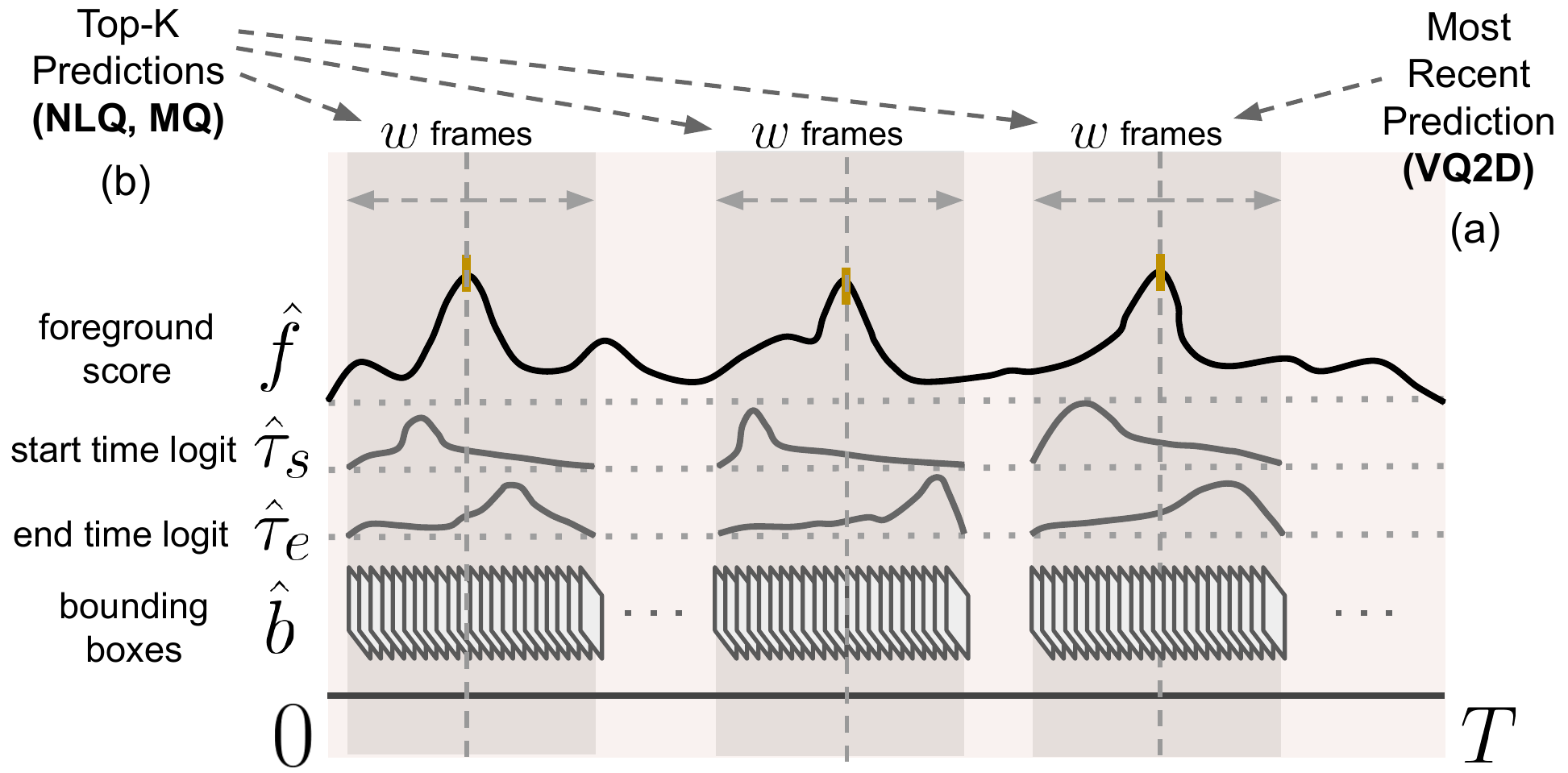}
    \vspace{-12pt}
   \caption{\textbf{Task-specific inference.} During inference, we identify all possible peaks from the time-series of foreground scores $\hat{f}$, and for generating VQ2D predictions (a), we retrieve the most recent occurrence of an object by forming predictions around the latest peak, while for MQ and NLQ (b), we use all the peaks for generating temporal segments and rank them according to their respective heights.}
   
   \label{fig:inference}
   \vspace{-10pt}
\end{figure}

\vspace{0.07in}
\noindent
{\bf Prediction heads.}
Similar to TubeDETR \cite{yang2022tubedetr}, we use MLPs to predict for each frame $t$: (a) the relative coordinates for the bounding box $\hat{\mathbf{b}}_t \in [0, 1]^{4}$ related to the query, (b) the score $\hat{\tau}_{s,t}$ that frame $t$ is the start of an answer to the query, and (c) the score  $\hat{\tau}_{e,t}$ that frame $t$ is the end of an answer. To better handle long-form videos, where multiple segments can be relevant to the query, we also output a confidence score $\hat{f}_t$ for each frame being relevant to the query. In summary, our model outputs 7-dimensional vector, $[\hat{\mathbf{b}}, \hat{\boldsymbol\tau}_s, \hat{\boldsymbol\tau}_e, \hat{\mathbf{f}}] \in \mathbb{R}^{T \times 7}$, as predictions for every frame in the video (shown in Fig \ref{fig:inference} (a)).

\vspace{0.07in}
\noindent
{\bf Task-specific inference.}
Due to the limitations of GPU memory our model cannot process all frames of a long-form video at once. To handle that, we adopt a sliding window approach, where we train using video segments of fixed duration of $w$ frames, and perform inference by aggregating model outputs in a sliding window fashion. These outputs then need to be post-processed to generate task-appropriate response tracks. To predict a spatio-temporal tube, we form probabilities for start and end time, $\hat{\mathbf{p}}_s$ and $\hat{\mathbf{p}}_e$ by taking {\tt softmax} over the logits $\hat{\boldsymbol\tau}_s$ and $\hat{\boldsymbol\tau}_e$ respectively. We infer start and end frame indices, $\hat{s}$ and $\hat{e}$, by choosing {\tt argmax} over the distributions such that $\hat{e} > \hat{s}$. The spatial predictions are formed by selecting bounding boxes in the predicted start/end range, resulting in space-time tube $\{ \hat{\mathbf{b}}_t \}_{t=\hat{s}}^{\hat{e}}$.

\vspace{0.07in}
\noindent
{\bf Inference: VQ2D.} The task requires finding the most recent occurrence of a query object. Therefore, we use foreground scores $\hat{\mathbf{f}} \in \mathbb{R}^T$ to identify a set of peaks in the scores which would indicate high probability of occurrence. From the set of peaks, we select the most recent one as our candidate peak as shown in Fig \ref{fig:inference}(a). We sample a window of size $w$ frames around the peak, and predict start and end frame indices, $\hat{s}$ and $\hat{e}$, by only considering the predictions within the window, \ie, outside the selected window, the start and end time probabilities, $\hat{p}_s$ and $\hat{p}_e$, are zero.

\vspace{0.07in}
\noindent
{\bf Inference: NLQ + MQ.} Since the two tasks require multiple predictions, we generate predictions using all identified peaks, and sort the predictions using the scores of the peaks. In addition NLQ and MQ don't require bounding boxes, so we only predict time intervals (or frame indices).

\input{tables/main-table-val-short}

\vspace{0.1in}
\noindent
{\bf Training.} 
During training, we randomly sample a window of $w$ frames (out of total $T$ frames) around a ground-truth temporal segment that answers the query (if there are multiple ground-truth segments we randomly pick one).
In the case that the ground-truth segment is longer than $w$, we randomly place the window covering a part of the ground-truth.

\vspace{0.1in}
\noindent
{\bf Losses.}
Our model is trained using \emph{spatial} losses, which refine the bounding box predictions, and \emph{temporal losses}, which refine the start/end/foreground frame predictions).
In particular, 
given ground-truth bounding boxes $\mathbf{b} \in [0,1]^{4 \times (e-s+1)}$ for every frame between the ground-truth start frame $s$, and end frame $e$, the spatial loss computes  $\mathcal{L}_1$ and generalized IoU ($\mathcal{L}_{gIoU}$) \cite{carion2020end} loss between $b$ and $\hat{b}$ as:
\begin{equation}
    \mathcal{L}_{spatial} = \lambda_{\mathcal{L}_1} \mathcal{L}_{\mathcal{L}_1} (\hat{\mathbf{b}}, \mathbf{b}) + \lambda_{gIoU} \mathcal{L}_{gIoU} (\hat{\mathbf{b}}, \mathbf{b}).
\end{equation}
For temporal losses, we define target start and end time distributions ($\mathbf{p}_s, \mathbf{p}_e \in [0,1]^{w}$) using standard Normal ($\mathcal{N}(., 1)$) centered at the ground-truth start and end frames, respectively. For predicted start and end time logits ($\hat{\boldsymbol\tau_s}, \hat{\boldsymbol\tau_e} \in \mathbb{R}^{w}$), we normalize them across $w$ frames using {\tt softmax} to get corresponding probabilities ($\hat{\mathbf{p}}_s, \hat{\mathbf{p}}_e \in [0,1]^w$), and compute KL-divergence ($\mathcal{L}_{KL}$) between target distribution $\mathbf{p}_s$ (or $\mathbf{p}_e$) and predicted distribution $\hat{\mathbf{p}}_s$ (or $\hat{\mathbf{p}}_e$). In addition, we also regress foreground scores, $\hat{\mathbf{f}} \in [0,1]^w$, which indicate the likelihood of a frame to be a part of the ground-truth. We form our target as $\mathbf{f} \in \{ 0, 1\}^w$, such that $\mathbf{p}$ is 1 within the start and end frame indices ($s$ and $e$), and 0 otherwise. We use positive-weighted binary cross-entropy loss ($\mathcal{L}_{BCE}$) between $\mathbf{f}$ and $\hat{\mathbf{f}}$ to compute foreground loss. Lastly, we have guided attention loss $\mathcal{L}_{att}(A)$ which promotes cross-attention weights in space-time decoder to be diagonal-like and have greater values in their corresponding temporal boundaries (see \cite{yang2022tubedetr}).
\begin{equation}
    \begin{split}
        \mathcal{L}_{temporal} = \lambda_{KL} & \left[\mathcal{L}_{KL} (\hat{\mathbf{p}}_s, \mathbf{p}_s) + \mathcal{L}_{KL} (\hat{\mathbf{p}}_e, \mathbf{p}_e) \right] \\
        &+ \lambda_f \mathcal{L}_{BCE} (\hat{\mathbf{f}}, \mathbf{f})
        +\lambda_{att} \mathcal{L}_{att} (A)
    \end{split}
\end{equation}

\vspace{-0.2cm}
Depending on the type of ground-truth annotations that are available for each task, the model is trained with different loss components.
Since, NLQ and MQ only have temporal annotations, 
we employ only the \emph{temporal} loss terms:
$\mathcal{L}_{NLQ} = \mathcal{L}_{MQ} = \mathcal{L}_{temporal}$. However, for VQ2D 
where we have ground-truth spatio-temporal tubes available, we apply all loss terms:
$\mathcal{L}_{VQ2D} = \mathcal{L}_{spatial} + \mathcal{L}_{temporal}$.

\vspace{0.08in}
\noindent
{\bf Multi-Task Learning.}
Accommodating the three Episodic Memory tasks into a single, unified architecture, has the potential for mutual beneficial cross-transfer of knowledge among the tasks. For joint-training we use Round-Robin Sampling \cite{lu202012} which samples batches from tasks one-by-one in a cyclical manner, effectively training the model with equal proportion of individual tasks even though the dataset sizes of the tasks could be different. The total loss in multi-task learning is the sum of the individual tasks' losses.

%% file: tables/main-table-val-short.tex
\begin{table*}[t]
\centering
\resizebox{\textwidth}{!}{%
\begin{tabular}{@{}ll|ccccccccccccc@{}}
\toprule
 &  & \multicolumn{4}{c}{VQ2D} & \multicolumn{4}{c}{NLQ} & \multicolumn{4}{c}{MQ} & \multicolumn{1}{l}{} \\ \midrule
 &  & \multirow{2}{*}{$\text{tAP}_{25}$} & \multirow{2}{*}{$\text{stAP}_{25}$} & \multirow{2}{*}{rec$\%$} & \multicolumn{1}{c|}{\multirow{2}{*}{Succ}} & \multicolumn{2}{c}{tIoU=0.3} & \multicolumn{2}{c|}{tIoU=0.5} & \multicolumn{2}{c}{tIoU=0.3} & \multicolumn{2}{c|}{tIoU=0.5} & \multirow{2}{*}{\begin{tabular}[c]{@{}c@{}}\# params\\ (\# models)\end{tabular}} \\ \cmidrule(l){7-14} 
 
 &  &  &  &  & \multicolumn{1}{c|}{} & R@1 & R@5 & R@1 & \multicolumn{1}{c|}{R@5} & R@1x & R@5x & R@1x & \multicolumn{1}{c|}{R@5x} & \multicolumn{1}{l}{} \\ \midrule

 \small\texttt{1} & Single-Task & 0.41 & 0.21 & 29.4 & \multicolumn{1}{c|}{\textbf{61.2}} & 6.53 & 17.22 & 3.61 & \multicolumn{1}{c|}{9.58} & 33.64 & 55.38 & 23.86 & \multicolumn{1}{c|}{39.48} & 432M (3) \\

 \small\texttt{2} & All-Tasks (AT) & 0.41 & 0.19 & 26.4 & \multicolumn{1}{c|}{60.2} & \textbf{7.74} & 21.14 & 4.78 & \multicolumn{1}{c|}{12.60} & \textbf{34.82} & \textbf{56.68} & \textbf{25.58} & \multicolumn{1}{c|}{42.30} & 186M (1) \\

 \small\texttt{3} & AT $\ \ \rightarrow$ Tasks & \textbf{0.42} & \textbf{0.22} & \textbf{29.4} & \multicolumn{1}{c|}{61.1} & 7.69 & \textbf{21.17} & \textbf{5.21} & \multicolumn{1}{c|}{\textbf{12.98}} & 33.82 & 56.61 & 25.21 & \multicolumn{1}{c|}{\bf 42.83} & 432M (3)\\
 \bottomrule
\end{tabular}%
}
\smallskip
\vspace{-8pt}
\caption{\textbf{Multi-task learning on validation set.} We observe that, 1) our All-Tasks (AT) model (row \texttt{2}) trained using multi-task learning outperforms individual Single-Task (row \texttt{1}) models on $9$ out of $12$ metrics across the three tasks, and 2) further fine-tuning the All-Tasks (AT) model to individual tasks generally boost the performance (row \texttt{1} v/s row \texttt{3}).} 
\label{tab:multi-task-val-short}
\vspace{-8pt}
\end{table*}

%% file: experiments.tex
\section{Experiments}
\label{sec:experiments}

\paragraph{Datasets.} 
\label{sec:datasets}
Ego4D is a large-scale egocentric video dataset with benchmarks that evaluate first-person visual understanding \cite{grauman2022ego4d}. We experiment with the three episodic memory tasks. 
\textbf{VQ2D} consists of roughly $22k$ video-query pairs. The mean length of spatio-temporal tubes is approx. $15$ frames (3 secs). 
\textbf{NLQ} consists of $19.2k$ video-query pairs where language queries are based on $13$ query templates. The mean and mode of temporal tube length in train set is approx. $54$ ($11$ secs) and $6$ ($1.2$ sec) frames respectively. 
\textbf{MQ} has $22.2k$ action instances across $2.5k$ videos with $110$ pre-definined activity classes. The mean and mode of temporal tube length in train set is approx. $220$ ($44$ secs) and $7$ ($1.4$ sec) frames respectively. 
All three tasks are split into train:val:test sets with 3:1:1 ratio.

\vspace{0.05in}
\noindent
{\bf Evaluation metrics.} 
\label{sec:eval-metrics}
For each task, we use the evaluation metrics defined in Ego4D~\cite{grauman2022ego4d}. For \textbf{VQ2D} we report \emph{temporal AP} ($\text{tAP}_{25}$) and \emph{spatio-temporal AP} ($\text{stAP}_{25}$) which measure the average-precision of predicted temporal and spatio-temporal extent of a tube with the ground-truth at IoU $=0.25$. We also have \emph{Recovery} ($\text{rec}\%$) which calculates $\%$ of frames in predicted tube where bounding box has at least $0.5$ IoU with the ground-truth, and \emph{Success} (Succ) which measures whether prediction has any overlap with ground-truth as $\%$ of samples where prediction has at least $0.05$ IoU with the ground-truth. \textbf{NLQ} is evaluated using \emph{recall@k, tIoU=m}, with $k \in \{1,5\}$ and $m \in \{0.3, 0.5\}$, which measures the percentage of samples where at least one of the top-$k$ predictions has temporal IoU of at least $m$ with the ground-truth. For \textbf{MQ}, we have \emph{recall@kx, tIoU=m}, with $k \in \{1,5\}$ and $m \in \{0.3, 0.5\}$, which measures $\%$ of predicted instances of class $x$ that have at least one prediction with tIoU greater than $m$ in the top-$k$ results.

\vspace{0.05in}
\noindent
{\bf Implementation details.}
\label{sec:implementation-details}
Following TubeDETR \cite{yang2022tubedetr}, we use a ResNet-101 \cite{he2016deep} as our visual backbone and RoBERTa \cite{liu2019roberta} as the language model.
We pre-process videos 
by resizing them with a shorter side size of 320, and sampling them at 5 fps. During training, we sample a window of length $w$, where $w$ is $200$, $400$ and $400$ frames, decoded at $5$, $1$ and $1$ fps, and downsampled with stride $1$, $5$ and $5$ for VQ2D, NLQ and MQ, respectively. The rest of our hyperparameters are adopted from  TubeDETR \cite{yang2022tubedetr}: $N_e=N_d=6, d=256, \lambda_{\mathcal{L}_1}=5, \lambda_{\mathcal{L}_{gIoU}}=2, \lambda_{KL}=10, \lambda_{att}=1, \lambda_{f}=2$. We initialize our weights from MDETR \cite{kamath2021mdetr} which is pretrained on Flickr30K \cite{plummer2015flickr30k}, MS COCO \cite{chen2015microsoft} and Visual Genome \cite{krishna2017visual}. We train our models on 16 GPUs with an effective batch size of $16$ videos for 25 epochs with a drop in learning rate at every $10^{th}$ epoch by a factor of 10. The final model is selected based on best performance on a small subset of validation data. %

\subsection{Main Results} 
\label{sec:results}

\paragraph{Single-task v/s multi-task.}
\label{sec:multi-task}
We train our model individually on the three tasks to setup a baseline for multi-task learning, Single-Task, see row \texttt{1} of Table \ref{tab:multi-task-val-short}. We use the same model architecture except in the case of VQ2D where text encoder is not needed. In addition, we train a single model on all the three tasks, All-Tasks (AT), shown in row \texttt{2}. For $9$ out of $12$ metrics across the three tasks, we obtain improvements in multi-task results over Single-task (row \texttt{1} v/s row \texttt{2}). 
Moreover, the total number of models required for all three tasks drops from $3$ to $1$, leading to $2.3 \times$ fewer parameters overall. 
We analyze additional pair-wise trained task combinations and transfer among tasks in Suppl Mat. 

We highlight that amount of data per task and the architecture is the same for Single-task and All-Tasks (AT) models; meaning, improvements are purely based on effective multi-task training and complementary information learned across tasks. 
All-Task (AT) model results in largest improvements, over the Single-Task counterpart, on NLQ task owning to significant synergies with VQ2D. In particular, since VQ2D focuses on inference of objects and places, the improvements in NLQ are largest on questions of those categories as well (see Table~\ref{tab:qanalysis}). Meanwhile, improvements on people questions over Single-Task model are negligible.

\input{tables/breakdown-nlq-perf}

We also experiment with fine-tuning All-Tasks (AT) model to individual tasks with the idea that AT model may allow downstream tasks to take advantage from multi-task joint pre-training \cite{lu202012}. Shown in row \texttt{3}, for $11$ out of $12$ metrics across the three tasks, we observe comparable or better performance than Single-Task (row \texttt{2}). 
The improvements of fine-tuning (row \texttt{3}) over the All-Tasks (AT) model (row \texttt{3}) are modest.
This highlights that our multi-task learning is effective and unified model is itself competitive. 

\begin{figure}[]
  \centering
   \includegraphics[width=0.98\linewidth]{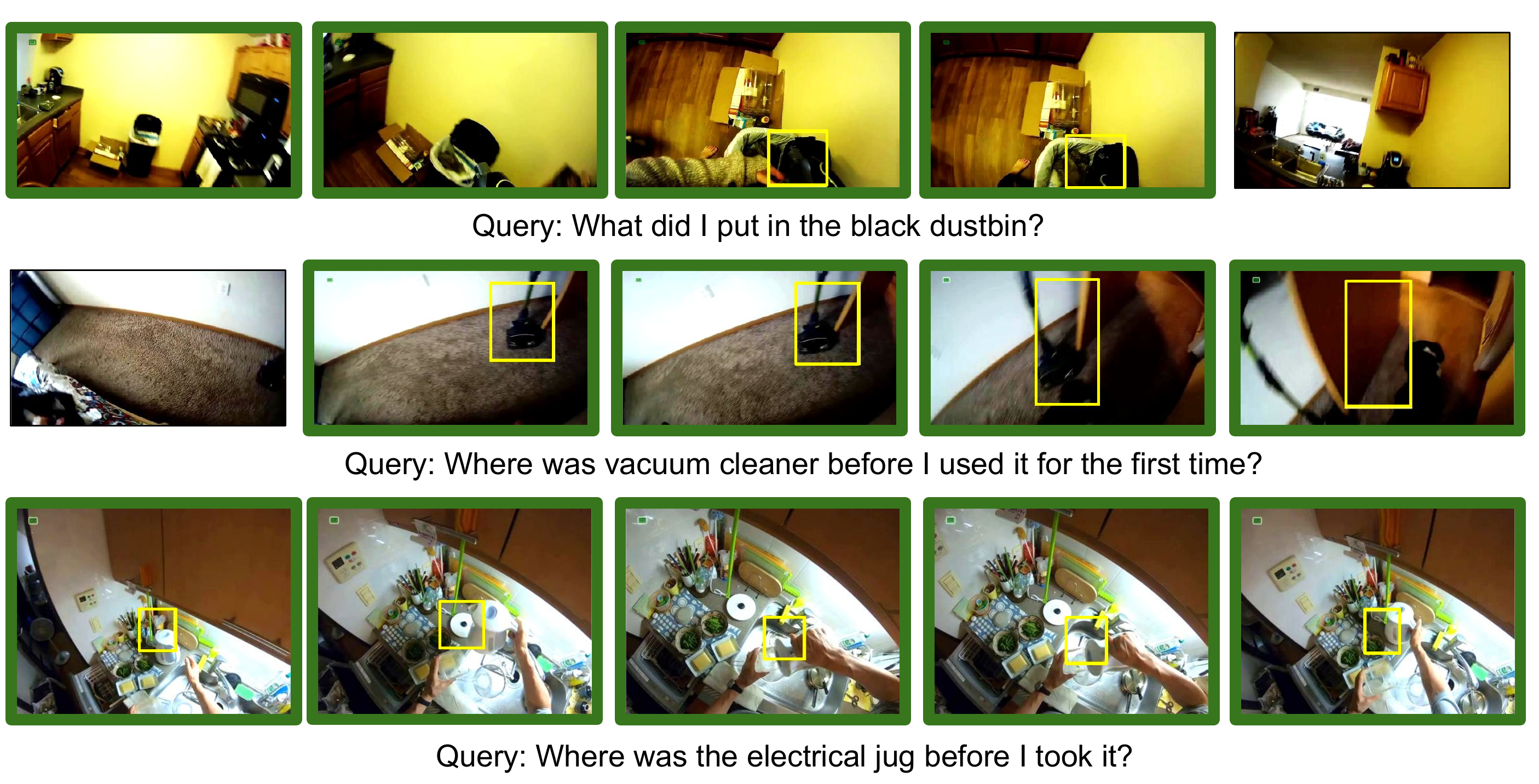}
   \vspace{-0.1in}
   \caption{\textbf{Spatio-temporal predictions for NLQ with the \emph{All-Tasks (AT)} model.} 
   Predictions are shown with \textcolor{Dandelion}{yellow} bounding boxes, while ground-truth temporal segments are denoted with \textcolor{OliveGreen}{green} borders. Although the model has not be trained with ground-truth spatio-temporal tubes for language queries, multi-task training results in some meaningful results for zero-shot spatio-temporal grounding.
}
   \label{fig:qualitative}
   \vspace{-4pt}
\end{figure}

\input{tables/nlq-spatial-zero-shot}
\input{tables/sota-table-cvpr22-updated}

\input{tables/ablation-mq-all}
\vspace{0.07in}
\noindent
{\bf Synergy across different tasks.}
Another benefit of training a unified model on multiple tasks is the ability to transfer learned concepts across the tasks. 
Specifically, for NLQ, even though the model was trained on (language query, temporal target) pairs, we can turn on the spatial prediction branch of the model, and make spatial predictions on language query, despite the fact that the spatial branch was trained only on the (visual query, spatio-temporal target) pairs in VQ2D. We call this setup \textit{zero-shot spatio-temporal grounding} where a unified model can transfer learned concepts across the modalities.

We benchmark the performance of spatial predictions by manually annotating a subset of NLQ data with spatial bounding boxes. 
Table \ref{tab:nlq-zero-shot} shows the results comparing Single-Task model on NLQ (``NLQ only") and All-Tasks (AT) trained on all tasks. We added two random baselines for spatial predictions - \textit{random boxes}: which produces random boxes for each frame, and \textit{random centered boxes}: which produces center-biased random boxes which are centered for each frame. We observe that, 1) our All-Tasks model outperforms Single-Task ``NLQ only" model on the temporal benchmark (similar conclusion as in Table \ref{tab:multi-task-val-short}), and 2) spatio-temporal predictions from All-Tasks model outperforms the two random baselines (repeated $5$ times). This shows that our All-Tasks model produces non-trivial zero-shot spatio-temporal localization on task for which it wasn't trained on. We also show a few examples of spatio-temporal predictions on NLQ task using All-Tasks model in Figure \ref{fig:qualitative}.

\subsection{Ablations}

We ablate different design choices using our All-Tasks (AT) model on MQ in Table \ref{tab:ablation-mq-all}. More ablation study results on VQ and NLQ are available in the Suppl Mat. 1) \textbf{w/o text encoder}: instead of training a language encoder for query “When did I do $<$activity c$>$?” , we experiment with one-hot encodings for each moment class and embed them using 1-layered MLP, 2) \textbf{w/o modality-specific time-embedding}: instead of having separate time embeddings for visual and text queries ($\mathbf{e}^{visual}, \mathbf{e}^{text}$), we report results on using a single time embedding, 3) \textbf{w/o round-robin sampling}: instead of uniformly sampling over three tasks during training, we show results for simply concatenating the datasets, 4) \textbf{w/o fine-tuning text encoder}: , we report results on freezing the text encoder instead of jointly fine-tuning it with the query objectives, 5) \textbf{w/o multi-scale inference}: instead of accumulating inference results at different scales (fps: $0.2, 0.5, 1, 1.66, 5$), we show the results with a single scale (fps: $5$), 6) \textbf{w/o overlapping windows}: increasing the stride between adjacent sliding windows to remove overlapping during inference. 
We observe significant drop in performance across all tested ablations, indicating the effectiveness of our technical designs in achieving a unified multi-task model.

\vspace{0.05in}
\noindent
We also study the impact of the foreground head (detailed in Section~\ref{sec:multi-task-learn-infer}) for generating predictions on long videos. Specifically for VQ2D, we leverage the foreground scores $\hat{\mathbf{f}}$ over a video to identify the most recent occurrence of a query object. This is in contrast to TubeDETR \cite{yang2022tubedetr} which makes predictions on whole-video without accounting for this bias. Our results show that using foreground predictions improves the baseline from 0.25 (0.15) to 0.41 (0.21) for tAP25 (stAP25).

\subsection{Comparison with Existing Works}
\label{sec:sota-comparison}
\noindent We compare our approach with prior works in Table \ref{tab:sota-table}. Note that we only report the methods trained on task-specific data (without additional large-scale pretraining on $3.85$M video-text Ego4D corpus) due to space limit and leave the others to the Suppl Mat.
We first observe that compared to baselines reported in Ego4D paper \cite{grauman2022ego4d} (rows \texttt{1} - \texttt{3}), our All-Tasks model (row \texttt{9}) outperforms on \textbf{8 out of 9} metrics across the three tasks. Second, compared to state-of-the-art on VQ2D task: Xu \textit{et al.} \cite{xu2022my} (row \texttt{4}), our All-Tasks model (row \texttt{9}) outperforms on $3$ out of $4$ metrics on VQ2D. Third, compared to other approaches which use different combinations of backbone architectures and pre-training data where EgoVLP \cite{lin2022egocentric} uses Frozen \cite{bain2021frozen} backbone with CC3M \cite{sharma2018conceptual} and WebVid2M \cite{bain2021frozen} pre-training, InternVideo \cite{chen2022internvideo} uses VideoMAE \cite{tong2022videomae} backbone with Kinetics-700 \cite{kay2017kinetics} pre-training, and ActionFormer \cite{zhang2022actionformer} uses SlowFast \cite{feichtenhofer2019slowfast} backbone with Kinetics-400 \cite{kay2017kinetics} pre-training; our All-Tasks (AT) model outperforms on NLQ (row \texttt{5} v/s \texttt{9}), and outperforms $2$ out of $3$ related works on MQ (rows \texttt{6} - \texttt{8} v/s row \texttt{9}), despite the fact that these approaches use dedicated architectures designed specifically for each individual task.

%% file: tables/breakdown-nlq-perf.tex
\begin{table}[]
\centering
\resizebox{\columnwidth}{!}{%
\begin{tabular}{@{}llccc@{}}
\toprule
                         &                                                & \multicolumn{3}{c}{R@$5$, tIoU=$0.5$}                                                                                       \\ \midrule
Category                 & Template                                       & \begin{tabular}[c]{@{}c@{}}NLQ\\ only\end{tabular} & All-Tasks & \begin{tabular}[c]{@{}c@{}}Gain\\ (in \%)\end{tabular} \\ \midrule
\multirow{9}{*}{Objects} & Where is object X before / after event Y?      & 6.21                                               & 7.30      & +17.50                                                   \\
                         & Where is object X?                             & 10.29                                              & 13.42     & +30.43                                                   \\
                         & What did I put in X?                           & 5.43                                               & 7.67      & +41.18                                                   \\
                         & How many X's? (quantity)                       & 17.67                                              & 23.67     & +33.96                                                   \\
                         & What X did I Y?                                & 9.94                                               & 13.78     & +38.71                                                   \\
                         & In what location did I see object X?           & 10.24                                              & 11.95     & +16.67                                                   \\
                         & What X is Y?                                   & 10.13                                              & 12.42     & +22.58                                                   \\
                         & \textbf{State of an object}                             & \textbf{11.31}                                              & \textbf{22.02}     & \textbf{+94.74}                                                   \\
                         & \textbf{Where is my object X?}                          & \textbf{6.49}                                               & \textbf{11.69}     & \textbf{+80.00}                                                   \\ \midrule
Place                    & Where did I put X?                             & 5.43                                               & 7.67      & +41.18                                                   \\ \midrule
\multirow{3}{*}{People}  & Who did I interact with when I did activity X? & 12.75                                              & 11.76     & -7.69                                                    \\
                         & Who did I talk to in location X?               & 15.66                                              & 16.87     & +7.69                                                    \\
                         & When did I interact with person with role X?   & 4.00                                               & 4.00      & 0.00                                                     \\ \bottomrule
\end{tabular}%
}
\caption{\textbf{All-Tasks v/s Single-Task on NLQ.} Performance on NLQ across $13$ question types. We observe that All-Tasks model brings larger improvement for object-centric questions compared to Single-Task (``NLQ only").}
\label{tab:qanalysis}
\vspace{-2pt}
\end{table}

%% file: tables/nlq-spatial-zero-shot.tex
\begin{table}[]
\centering
\resizebox{\columnwidth}{!}{%
\begin{tabular}{@{}c|cccc|c@{}}
\toprule
\multicolumn{1}{l|}{}           & \multicolumn{4}{c|}{Spatio-temporal}                                                                                          & Temporal                                                                                                                                             \\ \midrule
\multirow{2}{*}{Model}  & \multirow{2}{*}{\begin{tabular}[c]{@{}c@{}}spatial\\ branch\end{tabular}} & \multicolumn{2}{c}{stIoU=0.3} & \multirow{2}{*}{\begin{tabular}[c]{@{}c@{}}mean\\ stIoU\end{tabular}}      & \multirow{2}{*}{\begin{tabular}[c]{@{}c@{}}mean\\ tIoU\end{tabular}}   \\ 
\cmidrule(lr){3-4} 
&       & R@1      & R@5        &     &                         \\ \midrule
NLQ-only & N/A & - & - & - & 5.35 \\ \midrule
\multirow{4}{*}{\begin{tabular}[c]{@{}c@{}}MINOTAUR\\ (All-Tasks)\end{tabular}}& random boxes  & 0 $\pm$ 0         & 0 $\pm$ 0             & 0.40 $\pm$ 0.04 & \multirow{4}{*}{\textbf{8.35}} \\ 
& \begin{tabular}[c]{@{}c@{}}random centered\\ boxes\end{tabular}           & 0 $\pm$ 0         & 0.47 $\pm$ 0.38       & 1.25 $\pm$ 0.03 & \\
\cmidrule(l){2-5}
 & All-Tasks & \textbf{2.33}        & \textbf{4.65}            & \textbf{2.27} &  \\
\bottomrule
\end{tabular}%
}
\caption{\textbf{Zero-shot spatio-temporal grounding on NLQ.} We evaluate spatio-temporal predictions on NLQ by annotating a subset of validation videos ($=130$), and comparing it with random baselines. We observe non-trivial performance despite the model not trained on (language query, spatio-temporal target).}
\label{tab:nlq-zero-shot}
\vspace{-8pt}
\end{table}

%% file: tables/sota-table-cvpr22-updated.tex
\begin{table*}[ht!]
\centering
\resizebox{0.8\textwidth}{!}{%
\begin{tabular}{@{}ll|cccc|cccc|c@{}}
\toprule
 & & \multicolumn{4}{c|}{VQ2D} & \multicolumn{4}{c|}{NLQ} & \multicolumn{1}{c}{MQ} \\ \cmidrule(l){1-11} 
 & &  &  &  &  & \multicolumn{2}{c}{tIoU=0.3} & \multicolumn{2}{c|}{tIoU=0.5} & \multicolumn{1}{c}{tIoU=0.5} \\ \cmidrule(l){7-11} 
 & \multirow{-3}{*}{Methods} & \multirow{-2}{*}{$\text{tAP}_{25}$} & \multirow{-2}{*}{$\text{stAP}_{25}$} & \multirow{-2}{*}{rec$\%$} & \multirow{-2}{*}{Succ} & R@1 & R@5 & R@1 & R@5 & R@1x \\ \midrule
 \small\texttt{1} & Siam-RCNN \cite{grauman2022ego4d} & 0.21 & 0.13 & 34.0 & 41.6 & - & - & - & - & -  \\
 \small\texttt{2} & VSLNet \cite{grauman2022ego4d,zhang2020span} & - & - & - & - & 5.47 & 11.21 & 2.80 & 6.57 & - \\
 \small\texttt{3} & VSGN \cite{grauman2022ego4d,zhao2021video} & - & - & - & - & - & - & - & - & 24.25  \\ 
 \small\texttt{4} & Xu \textit{et al.} \cite{xu2022my} & 0.26 & 0.18 & \textbf{43.2} & 48.1 & - & - & - & - & - \\
 \small\texttt{5} & EgoVLP: VSLNet \cite{lin2022egocentric} & - & - & - & - & 4.87 & 8.67 & 2.50 & 4.97 & -  \\
\small\texttt{6} & EgoVLP: VSGN \cite{lin2022egocentric} & - & - & - & - & - & - & - & - & 19.74  \\ 
\small\texttt{7} & InternVideo \cite{chen2022internvideo} & - & - & - & - & - & - & - & - & \textbf{27.71} \\
\small\texttt{8} & ActionFormer \cite{mu2022actionformerego4d} & - & - & - & - & - & - & - & - & 24.25 \\ \midrule

\small\texttt{9} & Ours: All-Tasks (AT) & 0.41 & 0.19 & 26.5 & 60.6 & 7.47 & 19.61 & 4.85 & 12.09 & 24.62 \\ 
\small\texttt{10} & Ours: AT $\ \rightarrow \ $ Tasks & \textbf{0.42} & \textbf{0.21} & 29.0 & \textbf{61.4} & \textbf{8.07} & \textbf{20.63} & \textbf{5.52} & \textbf{12.94} & 24.44 \\

\bottomrule

\end{tabular}
}
\caption{\textbf{Comparison with existing works on test set.} We observe, 1) our approach outperforms Ego4D \cite{grauman2022ego4d} baselines (row \texttt{1} - \texttt{3}) on $8$ out of $9$ metrics across the three tasks, 2) our approach performs comparably to SOTA using task-specific data (without pretraining on additional $3.85$M Ego4D video-text corpus). For MQ, only R@1x at tIoU=$0.5$ is reported on test set.}
\vspace{-6pt}
\label{tab:sota-table}
\end{table*}

%% file: tables/ablation-mq-all.tex
\begin{table}[]
\centering
\resizebox{0.94\columnwidth}{!}{%
\begin{tabular}{@{}clcc@{}}
\toprule
\multicolumn{2}{c}{\multirow{2}{*}{MQ}}     & \multicolumn{2}{c}{tIoU=0.5} \\ \cmidrule(l){3-4} 
\multicolumn{2}{c}{}                        & R@1x          & R@5x         \\ \midrule
\multicolumn{2}{c}{MINOTAUR (All-Tasks model)}                                     & \textbf{25.58}         & \textbf{42.30}        \\ \midrule
\multirow{1}{*}{Input} & w/o text encoder   & 18.22         & 34.54        \\
 \midrule
& w/o modality-specific time embedding          & 24.39         & 42.04        \\
\multirow{1}{*}{Train} & w/o round-robin sampling            & 21.97         & 39.64        \\
 & w/o fine-tuning text encoder & 18.71         & 31.40        \\
 \midrule
\multirow{2}{*}{Inf.}& w/o multi-scale inference                     & 20.86         & 38.48        \\
& w/o overlapping windows                       & 20.01         & 37.31        \\ 
\bottomrule
\end{tabular}%
}
\caption{\textbf{Ablation study on MQ}. We ablate individual components in MINOTAUR by removing each independently.}
\label{tab:ablation-mq-all}
\end{table}

%% file: conclusion.tex
\section{Conclusion}
\vspace{-0.05in}
We present a unified approach for grounding multi-modal queries in long-videos with different degrees of spatio-temporal outputs. To achieve that, we propose a multi-task training and inference strategies that enables learning on data with varying degree of spatio-temporal annotations. Finally, we observe effectiveness of our approach over task-specific single architectures in terms of generalization and transfer of learned abilities such as zero-shot spatio-temporal grounding.

%% file: appendix.tex
\clearpage
\appendix

The supplementary is structured in the following way:
\begin{itemize}
\setlength\itemsep{-0.12em}
    \item Section \ref{sec:appendix-comparison-sota} details complete comparison to SOTA 
    \item Section \ref{sec:appendix-pair-wise-joint-transfer} discusses pair-wise joint and transfer learning results
    \item Section \ref{sec:appendix-zero-shot} discusses details of zero-shot spatio-temporal grounding on NLQ
    \item Section \ref{sec:appendix-semi-vos} presents results on a novel task: Semi-Supervised Video Object Segmentation 
    \item Section \ref{sec:appendix-implementation-details} discusses implementation details
    \item Section \ref{sec:appendix-ablation} presents additional ablation studies
    \item Section \ref{sec:appendix-qualitative-vq2d} shows qualitative results on VQ2D
    \item Section \ref{sec:appendix-qualitative-nlq} shows qualitative results on NLQ
\end{itemize}

\section{Comparison to Existing Work}
\label{sec:appendix-comparison-sota}
\input{tables/sota-table-cvpr22-updated-full}
As discussed in Section \textcolor{red}{4.3} and Table \textcolor{red}{4} of the main paper, we reported methods trained \textit{only} on the task-specific data, i.e., VQ2D, NLQ and MQ tasks of Ego4D \cite{grauman2022ego4d} for fair comparison. We did this to differentiate between approaches that use 1.) only task-specific data, such as our own, and 2.) task-specific data \textit{with} additional large-scale pre-training on 3.85M video-text Ego4D corpus, where the methods in the latter case are positioned to leverage additional data and perform better.
In this section, we provide details of the methods that perform large-scale pre-training and present results in Table \ref{tab:sota-table-full} for methods \textit{with} and \textit{without} additional large-scale pre-training.

\vspace{5pt}
\noindent
\textbf{Additional large-scale pre-training.} EgoVLP \cite{lin2022egocentric} proposes to use EgoCLIP dataset which is chosen from the broad Ego4D \cite{grauman2022ego4d} data and contains $3.85$M video-text pairs covering egocentric daily human activities, with the idea that pre-training on this additional data would benefit task-specific benchmarks. Indeed we observe notable improvements in performance (rows \texttt{11} - \texttt{15}), however, we note that our approach is trained only on the task-specific data (VQ2D, NLQ and MQ), which is orders of magnitude smaller than EgoCLIP ($60$K v/s $3.85$M). Pre-training with additional data also has the potential to benefit our model, but it's not the focus of this paper.

\input{tables/pair-wise-joint-transfer-results-val}
\section{Pair-wise joint training and transfer results.}
\label{sec:appendix-pair-wise-joint-transfer}
In the main paper, we presented joint-training results (\textit{All-Tasks} model) and compare it with \textit{Single-Task} models in Section \textcolor{red}{4.1} and Table \textcolor{red}{1}. Here, we present an exhaustive analysis of synergies among the tasks by exploring all pair-wise joint and transfer learning results in Table \ref{tab:all-joint-val}.

\vspace{4pt}
\noindent
\textbf{Pair-wise joint-training.} We explore all pair-wise combinations among the three tasks, shown in rows \texttt{3} - \texttt{5}. Compared to the Single-Task models on the same data in row \texttt{2}, we observe performance improvements for $2$ out of the $3$ tasks, namely, for NLQ and MQ; and for VQ2D the performance stays competitive to the Single-Task (row \texttt{1}) and decreases slightly w.r.t. spatio-temporal performance. Furthermore, we found performance of \textbf{NLQ} in the case of MQ + NLQ (row \texttt{3}) to be higher than NLQ + VQ2D (row \texttt{5}), which indicates greater synergy between NLQ and MQ tasks. On the other hand, we found performance of \textbf{MQ} in the case of MQ + VQ2D (row \texttt{4}) to be higher than MQ + NLQ (row \texttt{3}), which indicates greater synergy between MQ and VQ2D. We also found \textit{negative} transfer in the case of NLQ + VQ2D (row \texttt{5}), where both the tasks' performances reduce w.r.t. Single-Task (row \texttt{2}).

Moreover, the total number of models required for all three tasks drops to $2$ from $3$, since we can pick any two of the three pair-wise models, and the total number of parameters drops to approx $372$M ($2 \times 186$) from $432$M which is a factor of $1.16 \times$ drop.

\vspace{4pt}
\noindent
\textbf{Pair-wise transfer-training.} We also explore pair-wise transfer between the tasks where a model trained on one task is used as an initialization for the other task, shown in rows \texttt{7} - \texttt{12}. We observe that the transfer outperforms Single-Task performance in $4$ out of $6$ cases. For \textbf{MQ} (rows \texttt{7} - \texttt{8}), we observe better transfer performance from VQ2D compared to NLQ, which also confirms better synergy results between MQ and VQ2D from the above joint-training results. One possible explanation is that the queries in MQ are from a set of $110$ activity templates in the form of ``When did I do \textit{activity c}?", so its reliance on text encoder is low and sees more gain in performance from the visual task than its language counterpart. And for \textbf{NLQ} (rows \texttt{9} - \texttt{10}), we observe better transfer performance from MQ compared to VQ2D, which also confirms the negative transfer between NLQ and VQ2D in the above joint-training results. For VQ2D (rows \texttt{11} - \texttt{12}), we don't observe significant differences w.r.t. Single-Task performance (row \texttt{2}).

\section{Zero-shot spatio-temporal grounding}
\label{sec:appendix-zero-shot}
In the main paper (Section \textcolor{red}{4.1}), we presented evaluation of zero-shot spatio-temporal prediction capability of \textit{All-Tasks} model on NLQ task. Since, the NLQ data is annotated only with (language query, temporal target) pairs, we labelled spatial bounding-boxes for a subset of validation videos from NLQ, to serve as the benchmark for evaluating spatio-temporal predictions. We discuss the details of annotation below.

\vspace{4pt}
\noindent
\textbf{Selection of videos.} Given that we have $13$ question templates in the NLQ task, we uniformly sub-sample validation videos from each template. We randomly choose $10$ videos per template, amounting to $130$ videos in total (out of total $3.8$K validation data videos).

\input{tables/spatial-anno-template-wise}
\vspace{4pt}
\noindent
\textbf{Spatial annotation.}
Shown in Table \ref{tab:spatial-anno-template-wise}, we provide breakdown of the number of frames and duration of spatial bounding-boxes annotated per question template. We note that each question template is appropriately represented in the chosen subset, and there isn't any significant skewness in the data, where the mean and standard deviation of durations are $27.03$ and $3.97$ secs respectively. Overall, we annotated $1757$ frames in total at $5$ fps, which amounts to $351.4$ secs of spatially annotated video content, with the help of \textit{Make Sense}\footnote{\url{https://www.makesense.ai/}} web application.

\input{tables/nlq-spatial-zero-shot-full}
\vspace{4pt}
\noindent
\textbf{Evaluation.}
We use the standard evaluation protocol for NLQ, and extended it to account for spatial predictions as well, thereby extending the temporal-only metrics to spatio-temporal metrics. The complete table is shown in Table \ref{tab:nlq-zero-shot-full} where we also provide Recall scores for temporal-only evaluation. We note that despite evaluating on a small subset of validation videos, we observe trends consistent with the full validation set shown in Table \textcolor{red}{1} of the main paper, where the temporal-only performance of \textit{All-Tasks} model is better than \textit{Single-Task} model.

\begin{figure*}[]
  \centering
   \includegraphics[width=\linewidth]{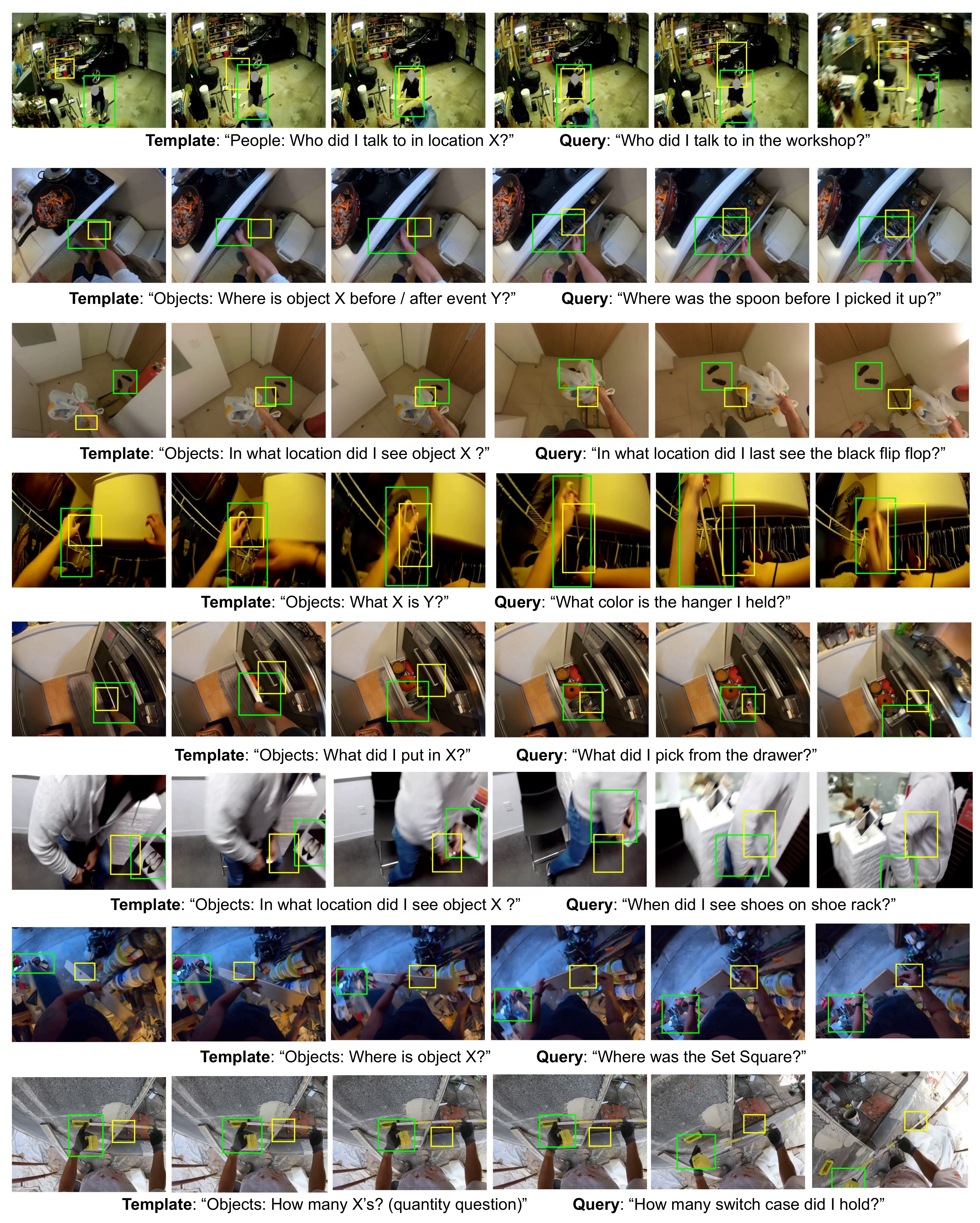}
   \vspace{-0.3in}
   \caption{\textbf{Zero-shot Spatio-temporal predictions for NLQ with the \emph{All-Tasks (AT)} model.} We show spatio-temporal predictions on a small subset of NLQ val set where the ground-truth bounding-boxes are annotated separately (not provided with the dataset). Predictions and ground-truth bounding boxes are shown with \textcolor{Yellow}{yellow} and \textcolor{OliveGreen}{green} borders respectively. Although the model has not been trained with ground-truth spatio-temporal tubes for language queries, multi-task training results in some meaningful results for zero-shot spatio-temporal grounding. The last two rows show failure cases.
}
   \label{fig:appendix-supp-zero-shot}
\end{figure*}

\vspace{4pt}
\noindent
\textbf{Qualitative predictions.}
In Figure \ref{fig:appendix-supp-zero-shot}, we show a handful of example spatial predictions and ground-truth on the validation subset of NLQ task. We note that to show zero-shot spatial prediction capability of the model, we plot spatial predictions on the frames belonging to ground-truth regardless of temporal prediction.

\section{Semi-supervised Video Object Segmentation (Semi-VOS)}
\label{sec:appendix-semi-vos}
In this section, we experiment with an additional benchmark task in egocentric domain to test extensibility of MINOTAUR to additional video structured prediction tasks, and generalizability of the features learned by \textit{All-Tasks} model on egocentric data domain. We choose Semi-Supervised Video Object Segmentation (Semi-VOS) task for this purpose. 

\vspace{4pt}
\noindent
\textbf{Task specification.} Given a video sequence and a ground-truth object mask in the first frame of the sequence, the Semi-VOS task requires segmenting the ground-truth object mask in the following frames of the sequence \cite{pont20172017}. We selected VISOR \cite{VISOR2022} as our benchmark dataset, as it is based on egocentric data \cite{damen2018scaling}, which is consistent with the data domain of Ego4D \cite{grauman2022ego4d}.

\vspace{4pt}
\noindent
\textbf{Approach.} We follow the proposed architecture with an additional prediction head in the form of a segmentation head provided by DETR \cite{carion2020end}. The only difference from prior prediction heads for foreground, start and end time and spatial bounding box is that the segmentation head uses multi-scale image features to generate segmentation masks, and it readily extends using our MINOTAUR framework.

\vspace{4pt}
\noindent
\textbf{Evaluation metrics.} Following standard protocols in DAVIS \cite{pont20172017,VISOR2022}, we use Jaccard Index/Intersection over Union ($\mathcal{J}$) and Boundary F-Measure ($\mathcal{F}$) metrics.

\input{tables/vos}
\vspace{4pt}
\noindent
\textbf{Comparison with baselines.} We report results in Table \ref{tab:vos}. First, we compare with Space-Time Memory Networks (STM) \cite{oh2019video} baseline reported in VISOR \cite{VISOR2022}. We observe that our method (\textit{All-Tasks}) performs comparably to STM \cite{oh2019video} which is pre-trained on COCO segmentation masks, while our method has not been pre-trained on any segmentation data. Second, our Ego4D (\textit{All-Tasks}) initialization outperforms MDETR \cite{kamath2021mdetr} initialization, which shows the generic and expressive nature of our features for other tasks at least in egocentric domain.

\begin{figure*}[]
  \centering
   \includegraphics[width=0.92\linewidth]{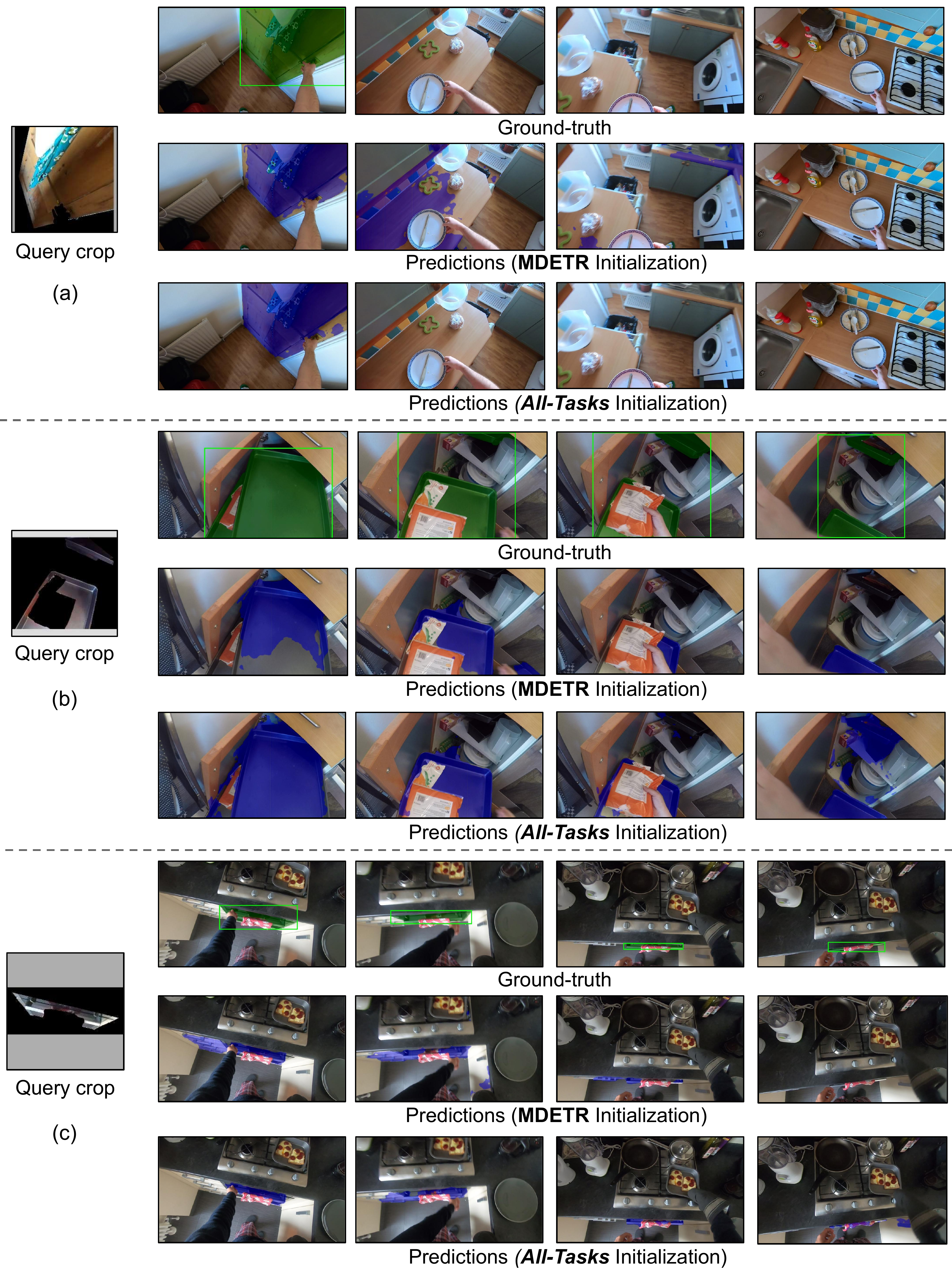}
   \vspace{-0.15in}
   \caption{\textbf{Additional results on Semi-Supervised Video Object Segmentation (Semi-VOS) task on VISOR \cite{VISOR2022} dataset.} Given a visual crop of an object \textit{mask} as query (shown in left), the Semi-VOS task requires outputs as segmentation masks of the object in the video sequence (shown as Ground-truth). We train MINOTAUR with an additional segmentation head \cite{carion2020end} on the task with 1.) MDETR \cite{kamath2021mdetr} and 2.) \textit{All-Tasks} initializations, and found the latter to perform better (with quantitative results shown in Table \ref{tab:vos}).}
   \label{fig:appendix-vos}
\end{figure*}

\vspace{4pt}
\noindent
\textbf{Example predictions.} We show a couple of example predictions along with the ground-truth in Figure \ref{fig:appendix-vos}, where we compare models trained with MDETR \cite{kamath2021mdetr} and \textit{All-Tasks} initialization.

\section{Implementation details}
\label{sec:appendix-implementation-details}

\vspace{4pt}
\noindent
\textbf{Identification of peaks.} With reference to Task-specific inference in the main paper (Section \textcolor{red}{3.1} and Figure \textcolor{red}{3}), during inference, we choose peaks by simple thresholding over the foreground scores $\hat{\mathbf{f}} \in \mathbb{R}^T$, where $T$ is the number of frames in a video. Following Siam-RCNN \cite{grauman2022ego4d}, we first apply a median filter, \texttt{medfilt} \footnote{\url{https://docs.scipy.org/doc/scipy/reference/generated/scipy.signal.medfilt.html}}, with kernel size of $5$. For VQ2D, we select the peaks with scores above $0.5$ threshold and pick the latest peak among them to form predictions. For MQ and NLQ, since we need to retrieve multiple predictions, we select all the peaks above $0.1$ threshold (maximum $1000$ in total), and rank them according to their scores.

\vspace{4pt}
\noindent
\textbf{Forming predictions from peaks.} As mentioned in the main paper (Section \textcolor{red}{3.1}), upon selection of the peak(s), we form predictions by sampling a window size of $w$ frames centered around every peak, and predict start and end frame, $\hat{s}$ and $\hat{e}$, by considering only the predictions within the window. We do this by masking out logits $\hat{\boldsymbol\tau}_s$ and $\hat{\boldsymbol\tau}_e$ outside the window. We then form start and end probabilities, $\hat{\mathbf{p}}_s, \hat{\mathbf{p}}_e \in \mathbb{R}^w$ specific to the window by taking \texttt{softmax}, and picking $\hat{s}, \hat{e}$ by taking \texttt{argmax} over the joint probability such that $\hat{e} > \hat{s}$.

For \textbf{VQ2D}, we choose a window of size $70$ frames centered around the chosen peak to form prediction. For \textbf{NLQ} and \textbf{MQ}, we perform \emph{test-time augmentation} by accumulating predictions at different frames-per-second (fps) to capture longer extents. For NLQ, we accumulate at fps $\{1,5 \}$, and, for MQ, we accumulate at fps: $\{0.2, 0.5, 1, 1.66, 5\}$, which we empirically found to perform best for validation set. We use NMS \cite{ren2015faster} with the threshold $0.4$ to disambiguate repeated or heavily overlapping entries.

\vspace{4pt}
\noindent
\textbf{Optimization details.} In addition to the implementation details discussed in the main paper (Section \textcolor{red}{4}), we use AdamW as the optimizer with learning rate for visual backbone as $1e^{-5}$, text encoder $5e^{-5}$ and the remaining components $5e^{-5}$. We train models for $25$ epochs with learning rate drop by a factor of $10$ at every $10^{th}$ epoch.

\section{Additional ablations}
\label{sec:appendix-ablation}

\input{tables/ablation-sampling-strategy}
\vspace{4pt}
\noindent
\textbf{Batch-level sampling: Round Robin v/s Concat.} In multi-task training, we can combine different tasks, by either concatenating the datasets together (\emph{Concat}), or we can sample batches in \emph{Round-Robin} fashion such that each dataset is sampled equally even though the individual dataset sizes are different. As shown in Table \ref{tab:ablation-sampling-strategy}, we found that \emph{Round-Robin} performs better on $10$ out of $12$ metrics across the three tasks.

\vspace{4pt}
\noindent
\textbf{MQ: fps 5 v/s fps 1.} As mentioned earlier that we train using fps 5 for VQ2D and fps 1 for NLQ and MQ. We show results for fps 5 on MQ in Table \ref{tab:ablation-fps-5-mq}. We found fps 1 to perform better, especially for MQ, since lower fps enables capturing longer temporal extents during training, and out of the three tasks MQ has the longest ground truth temporal extents, followed by NLQ and then VQ2D.

\input{tables/ablation-fps-5-mq}

\vspace{4pt}
\noindent
\textbf{Step size of sliding windows.} During evaluation, we accumulate predictions on long videos in a sliding-window fashion, where a window of size $w$ frames is used to gather predictions with step size of $k_{step}$ frames. In case of overlap between consecutive windows (or $k_{step} < w$), we average the predictions, $[\hat{\mathbf{b}}, \hat{\boldsymbol\tau}_s, \hat{\boldsymbol\tau}_e, \hat{\mathbf{f}}] \in \mathbb{R}^{T \times 7}$, across overlapping frames. We show how different step size $k_{step}$ affects performance in Table \ref{tab:ablation-step-size-mq}. In general, accumulating predictions with lower step size leads to better performance.

\input{tables/ablation-step-size-mq}

\section{Additional Qualitative Results on VQ2D}
\label{sec:appendix-qualitative-vq2d}
\begin{figure*}[]
  \centering
   \includegraphics[width=\linewidth]{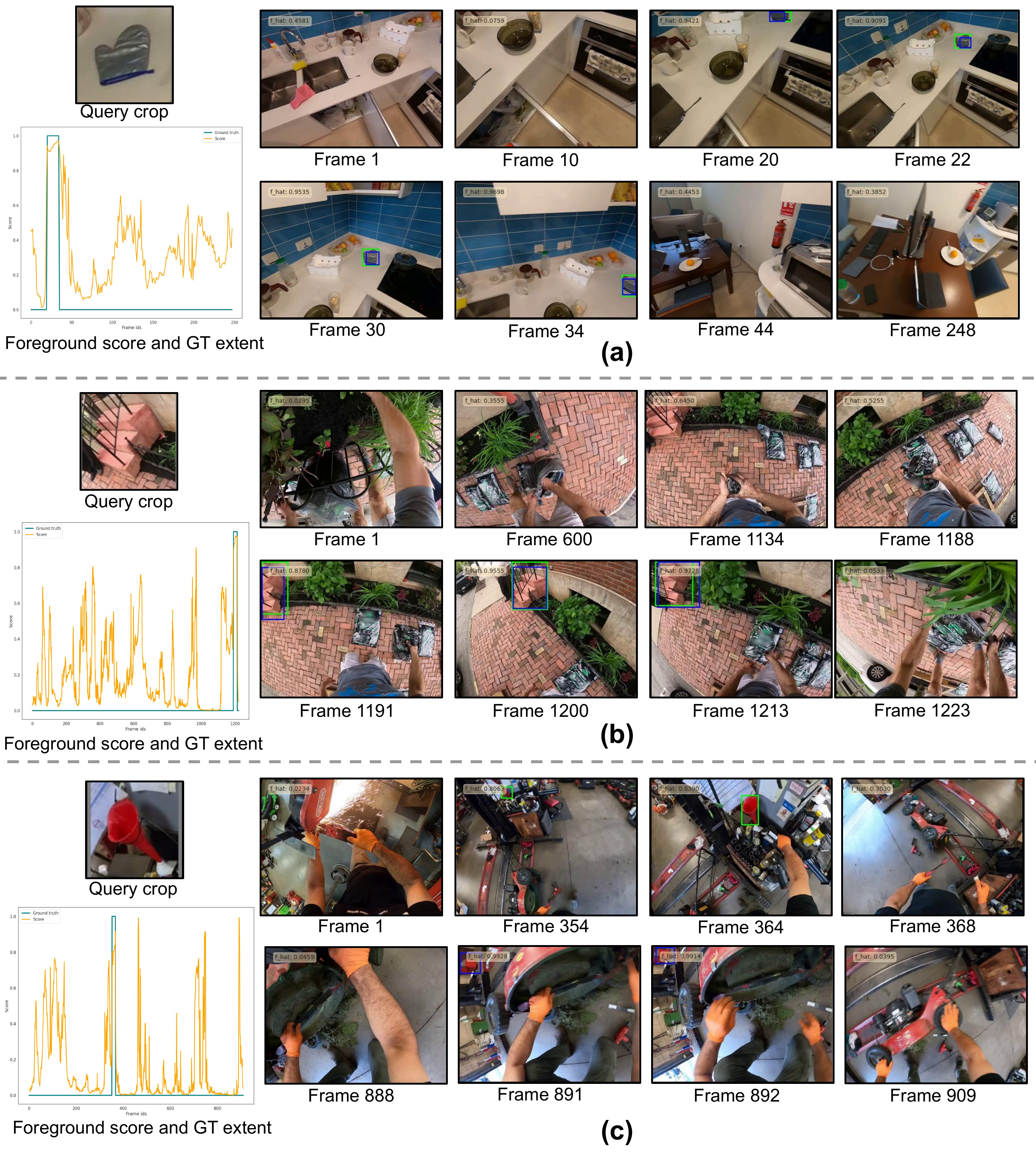}
   \vspace{-0.3in}
   \caption{\textbf{Qualitative results on VQ2D.} Given a visual crop of an object, the task requires localizing the \textit{most recent occurrence} of the object. For each example, we show query crop, plot of foreground score ($\mathbf{\hat{f}}$) (in \textcolor{yellow}{yellow}) and ground-truth extent (in \textcolor{OliveGreen}{green}), and frames with predicted (in \textcolor{blue}{blue}) and ground-truth bounding boxes (in \textcolor{green}{green}). In \textbf{(a)}, with a video of relatively short duration ($248$ frames), the model is able to correctly predict the object which is towards the beginning of the video with ground-truth frames: $[20, 34]$ and predicted frames: $[19,37]$, which can also be seen in the foreground score plot. In \textbf{(b)}, the video is of longer duration ($1223$ frames), and the model is able to correctly identify the object with ground-truth frames: $[1191, 1213]$ and predicted frames: $[1189,1215]$. However, note the appearance of the object before the ground-truth in frame $1134$ which can also be seen as a peak in the foreground score plot, but the model made prediction using the \textit{latest peak}. Lastly, \textbf{(c)} shows a failure case ($909$ frames), where the model incorrectly identifies a ``red looking" object as the prediction.
   }
   \label{fig:appendix-vq2d}
\end{figure*}

Please find the qualitative results in Figure \ref{fig:appendix-vq2d}.

\section{Additional Qualitative Results on NLQ}
\label{sec:appendix-qualitative-nlq}
\begin{figure*}[]
  \centering
   \includegraphics[width=\linewidth]{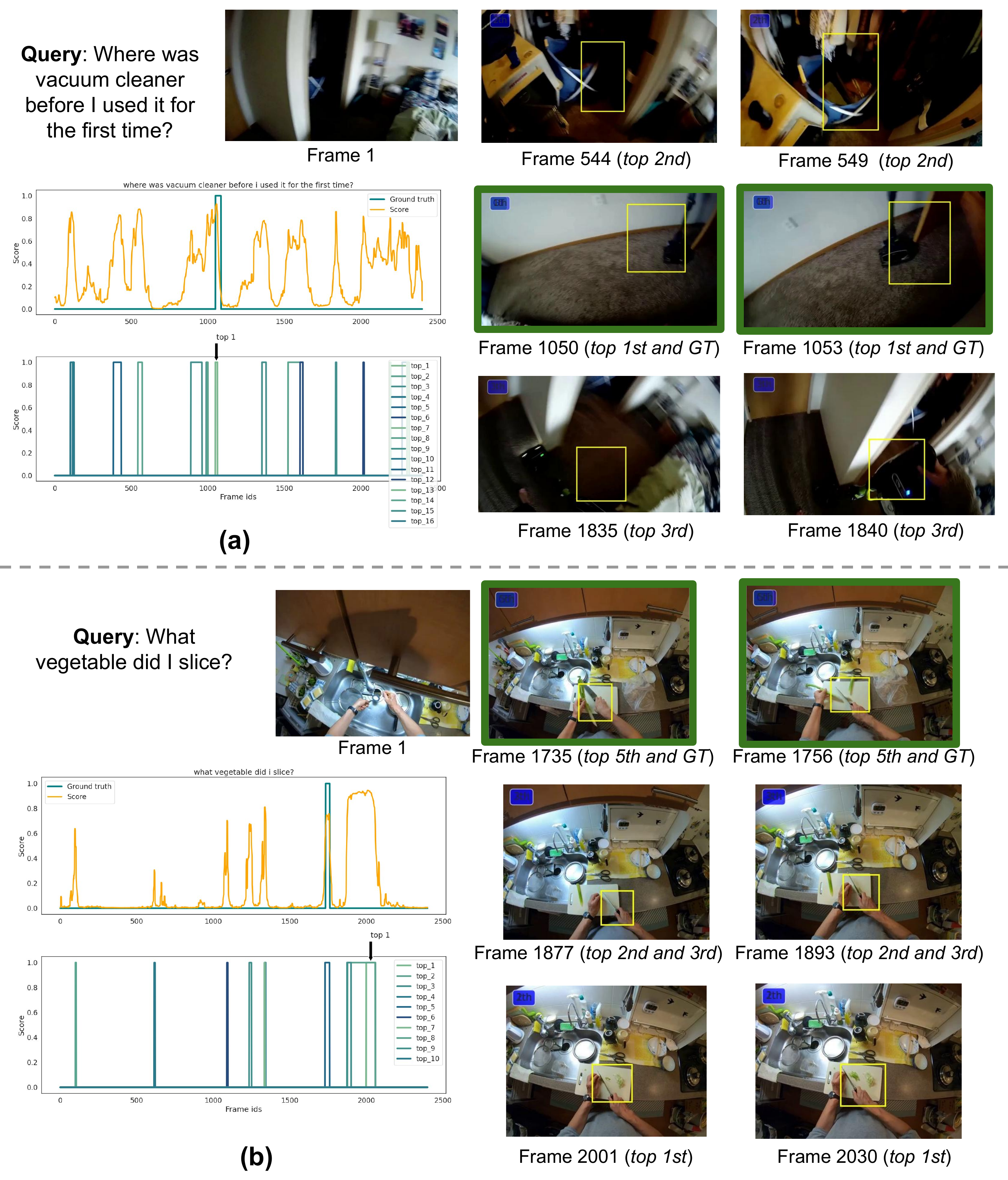}
   \vspace{-0.3in}
   \caption{\textbf{Qualitative results on NLQ.} Given a language query, the task requires temporally localizing the extent of the video where the answer can be found. For each example, we show language query, plot of foreground score ($\mathbf{\hat{f}}$) (in \textcolor{yellow}{yellow}) and ground-truth extent (in \textcolor{OliveGreen}{green}), and top-$k$ predictions; and frames with predicted bounding boxes (in \textcolor{yellow}{yellow}) and ground-truth temporal extent (in \textcolor{green}{green}). We note that the model is able to predict spatial bounding boxes, athough the training data for (language query, spatio-temporal target) was not provided. In \textbf{(a)}, we show that the top-$1st$ prediction (frames: $[1049,1063]$) and ground-truth (frames: $[1050,1085]$) extents overlap with each other. We also show top-$2nd$ (frames: $[544,571]$) and top-$3rd$ (frames: $[1835,1840]$) predictions as well. In \textbf{(b)}, we show that the top-$5th$ prediction (frames: $[1731,1760]$) and ground-truth (frames: $[1735,1762]$) extents overlap with each other. We also show top-$1st$ (frames: $[2000,2059]$), top-$2nd$ (frames: $[1877,2059]$), and top-$3rd$ (frames: $[1874,1900]$) predictions as well. Notice that we have overlapping predictions owing to the nature of the tasks and multi-scale predictions.
   }
   \label{fig:appendix-nlq}
\end{figure*}

Please find the qualitative results in Figure \ref{fig:appendix-nlq}.

%% file: tables/sota-table-cvpr22-updated-full.tex
\begin{table*}[]
\centering
\resizebox{0.95\textwidth}{!}{%
\begin{tabular}{@{}c|ll|cccc|cccc|c@{}}
\toprule
 & & & \multicolumn{4}{c|}{VQ2D} & \multicolumn{4}{c|}{NLQ} & \multicolumn{1}{c}{MQ} \\ \cmidrule(l){1-12} 
 & & &  &  &  &  & \multicolumn{2}{c}{tIoU=0.3} & \multicolumn{2}{c|}{tIoU=0.5} & \multicolumn{1}{c}{tIoU=0.5} \\ \cmidrule(l){8-12} 
 & & \multirow{-3}{*}{Methods} & \multirow{-2}{*}{$\text{tAP}_{25}$} & \multirow{-2}{*}{$\text{stAP}_{25}$} & \multirow{-2}{*}{rec$\%$} & \multirow{-2}{*}{Succ} & R@1 & R@5 & R@1 & R@5 & R@1x \\ \midrule
 & \small\texttt{1} & Siam-RCNN \cite{grauman2022ego4d} & 0.21 & 0.13 & 34.0 & 41.6 & - & - & - & - & -  \\
 \multirow{9}{*}{ \begin{tabular}[c]{@{}c@{}}\textbf{without}\\additional\\large-scale\\pre-training\end{tabular}} & \small\texttt{2} & VSLNet \cite{grauman2022ego4d,zhang2020span} & - & - & - & - & 5.47 & 11.21 & 2.80 & 6.57 & - \\
 & \small\texttt{3} & VSGN \cite{grauman2022ego4d,zhao2021video} & - & - & - & - & - & - & - & - & 24.25  \\ 
 & \small\texttt{4} & Xu \textit{et al.} \cite{xu2022my} & 0.26 & 0.18 & \textbf{43.2} & 48.1 & - & - & - & - & - \\
 & \small\texttt{5} & EgoVLP: VSLNet \cite{lin2022egocentric} & - & - & - & - & 4.87 & 8.67 & 2.50 & 4.97 & -  \\
& \small\texttt{6} & EgoVLP: VSGN \cite{lin2022egocentric} & - & - & - & - & - & - & - & - & 19.74  \\ 
& \small\texttt{7} & InternVideo \cite{chen2022internvideo} & - & - & - & - & - & - & - & - & \textbf{27.71} \\
& \small\texttt{8} & ActionFormer \cite{mu2022actionformerego4d} & - & - & - & - & - & - & - & - & 24.25 \\ \cmidrule(l){2-12}

& \small\texttt{9} & Ours: All-Tasks (AT) & 0.41 & 0.19 & 26.5 & 60.6 & 7.47 & 19.61 & 4.85 & 12.09 & 24.62 \\ 
& \small\texttt{10} & Ours: AT $\ \rightarrow \ $ Tasks & \textbf{0.42} & \textbf{0.21} & 29.0 & \textbf{61.4} & \textbf{8.07} & \textbf{20.63} & \textbf{5.52} & \textbf{12.94} & 24.44 \\
\midrule 

\multirow{5}{*}{ \begin{tabular}[c]{@{}c@{}}\textbf{with}\\additional\\large-scale\\pre-training\end{tabular}} & \small\texttt{11} & EgoVLP: VSLNet \cite{lin2022egocentric} & - & - & - & - & 10.46 & 16.76 & 6.24 & 11.29 & -  \\
& \small\texttt{12} & EgoVLP: VSGN \cite{lin2022egocentric} & - & - & - & - & - & - & - & - & 28.03  \\ 
& \small\texttt{13} & InternVideo \cite{chen2022internvideo} & - & - & - & - & 16.45 & 22.95 & 10.06 & 16.10 & -  \\
& \small\texttt{14} & InternVideo \cite{chen2022internvideo} & - & - & - & - & - & - & - & - & 41.13  \\
& \small\texttt{15} & ActionFormer \cite{mu2022actionformerego4d} & - & - & - & - & - & - & - & - & 42.54 \\
\bottomrule

\end{tabular}
}
\caption{\textbf{Comparison with existing works on test set \textit{with} and \textit{without} additional large scale pre-training.} In the main paper (Table \textcolor{red}{4}), we presented apples-to-apples comparison between MINOTAUR and related works using the same \textit{task-only} data (i.e., VQ2D, NLQ and MQ) without additional large-scale pre-training. Here, we show results on related approaches that leverage additional large-scale pre-training on $3.85$M Ego4D video-text corpus (EgoCLIP) to obtain better performance. Although, pre-training with additional data has the potential to benefit our model too, but it’s not the focus of this paper.}

\label{tab:sota-table-full}
\end{table*}

%% file: tables/pair-wise-joint-transfer-results-val.tex
\begin{table*}[t]
\centering
\resizebox{\textwidth}{!}{%
\begin{tabular}{@{}l|ll|ccccccccccccc@{}}
\toprule
 &  &  & \multicolumn{4}{c}{VQ2D} & \multicolumn{4}{c}{NLQ} & \multicolumn{4}{c}{MQ} & \multicolumn{1}{l}{} \\ \midrule
 &  &  & \multirow{2}{*}{$\text{tAP}_{25}$} & \multirow{2}{*}{$\text{stAP}_{25}$} & \multirow{2}{*}{rec$\%$} & \multicolumn{1}{c|}{\multirow{2}{*}{Succ}} & \multicolumn{2}{c}{tIoU=0.3} & \multicolumn{2}{c|}{tIoU=0.5} & \multicolumn{2}{c}{tIoU=0.3} & \multicolumn{2}{c|}{tIoU=0.5} & \begin{tabular}[c]{@{}c@{}}\# params\\ (\# models)\end{tabular} \\ \cmidrule(l){8-16} 
 &  &  &  &  &  & \multicolumn{1}{c|}{} & R@1 & R@5 & R@1 & \multicolumn{1}{c|}{R@5} & R@1x & R@5x & R@1x & \multicolumn{1}{c|}{R@5x} & \multicolumn{1}{l}{} \\ \midrule
\multicolumn{1}{c|}{\multirow{9}{*}{\rotatebox{90}{Joint}}} & \small\texttt{1} & Ego4D \cite{grauman2022ego4d} & 0.20 & 0.12 & \textbf{32.2} & \multicolumn{1}{c|}{39.8} & 5.45 & 10.74 & 3.12 & \multicolumn{1}{c|}{6.63} & 33.45 & \textbf{58.43} & 25.16 & \multicolumn{1}{c|}{\textbf{46.18}} & \multicolumn{1}{l}{} \\
\multicolumn{1}{c|}{} & \small\texttt{2} & Single-Task & 0.41 & 0.21 & 29.4 & \multicolumn{1}{c|}{\textbf{61.2}} & 6.53 & 17.22 & 3.61 & \multicolumn{1}{c|}{9.58} & 33.64 & 55.38 & 23.86 & \multicolumn{1}{c|}{39.48} & 432M (3) \\ \midrule
\multicolumn{1}{c|}{} & \small\texttt{3} & MQ + NLQ & - & - & - & \multicolumn{1}{c|}{-} & \textbf{7.72} & 19.80 & 4.75 & \multicolumn{1}{c|}{11.82} & 32.68 & 54.16 & 24.27 & \multicolumn{1}{c|}{41.41} & 185M (1) \\
\multicolumn{1}{c|}{} & \small\texttt{4} & MQ + VQ2D & 0.41 & 0.20 & 27.7 & \multicolumn{1}{c|}{60.2} & - & - & - & \multicolumn{1}{c|}{-} & 33.87 & 56.35 & 25.47 & \multicolumn{1}{c|}{42.32} & 186M (1) \\
\multicolumn{1}{c|}{} & \small\texttt{5} & NLQ + VQ2D & 0.40 & 0.19 & 27.7 & \multicolumn{1}{c|}{60.0} & 6.17 & 16.86 & 3.67 & \multicolumn{1}{c|}{10.12} & - & - & - & \multicolumn{1}{c|}{-} & 186M (1) \\
\multicolumn{1}{c|}{} & \small\texttt{6} & All-Tasks (AT) & 0.41 & 0.19 & 26.4 & \multicolumn{1}{c|}{60.2} & 7.74 & 21.14 & 4.78 & \multicolumn{1}{c|}{12.60} & 34.82 & 56.68 & 25.58 & \multicolumn{1}{c|}{42.30} & 186M (1) \\ \midrule

\multicolumn{1}{c|}{\multirow{9}{*}{\rotatebox{90}{Transfer}}} & \small\texttt{7} & NLQ $\ \, \, \rightarrow$ MQ & - & - & - & \multicolumn{1}{c|}{-} & - & - & - & \multicolumn{1}{c|}{-} & 32.70 & 55.17 & 23.67 & \multicolumn{1}{c|}{40.78} & \multirow{2}{*}{185M (1)} \\
 & \small\texttt{8} & VQ2D $\rightarrow$ MQ & - & - & - & \multicolumn{1}{c|}{-} & - & - & - & \multicolumn{1}{c|}{-} & \textbf{35.75} & 56.17 & \textbf{26.05} & \multicolumn{1}{c|}{41.78} &  \\ \cmidrule(l){2-16} 
 & \small\texttt{9} & MQ $\quad \rightarrow$ NLQ & - & - & - & \multicolumn{1}{c|}{-} & 6.94 & 19.93 & 4.13 & \multicolumn{1}{c|}{11.72} & - & - & - & \multicolumn{1}{c|}{-} & \multirow{2}{*}{185M (1)} \\
 & \small\texttt{10} & VQ2D $\rightarrow$ NLQ & - & - & - & \multicolumn{1}{c|}{-} & 6.69 & 17.73 & 4.26 & \multicolumn{1}{c|}{10.66} & - & - & - & \multicolumn{1}{c|}{-} &  \\ \cmidrule(l){2-16}
 & \small\texttt{11} & MQ $\ \, \rightarrow$ VQ2D & 0.40 & 0.21 & 29.1 & \multicolumn{1}{c|}{60.4} & - & - & - & \multicolumn{1}{c|}{-} & - & - & - & \multicolumn{1}{c|}{-} & \multirow{2}{*}{61M (1)} \\
 & \small\texttt{12} & NLQ $\rightarrow$ VQ2D & 0.42  & 0.20 & 29.2 & \multicolumn{1}{c|}{61.1} & - & - & - & \multicolumn{1}{c|}{-} & - & - & - & \multicolumn{1}{c|}{-} &  \\ \cmidrule(l){2-16}
 & \small\texttt{13} & AT $\ \ \rightarrow$ Tasks & \textbf{0.42} & \textbf{0.22} & 29.4 & \multicolumn{1}{c|}{61.1} & 7.69 & \textbf{21.17} & \textbf{5.21} & \multicolumn{1}{c|}{\textbf{12.98}} & 33.82 & 56.61 & 25.21 & \multicolumn{1}{c|}{42.83} & 432M (3)\\

 \bottomrule
\end{tabular}%
}
\smallskip
\vspace{-8pt}
\caption{\textbf{Pair-wise joint and transfer learning results on validation set.} We explore pair-wise relations between the tasks and found better synergy between MQ and VQ2D compared to MQ and NLQ, by observing both joint (row \texttt{3} vs \texttt{4}) and transfer (row \texttt{7} vs \texttt{8}) results. And similarly, we found better synergy between NLQ and MQ compared to NLQ and VQ2D using both joint (row \texttt{3} vs \texttt{5}) and transfer (row \texttt{9} vs \texttt{10}) results. We also observe \textit{negative} synergy between NLQ and VQ2D (row \texttt{5}) where joint-training leads both the tasks to underperform w.r.t. Single-Task (row \texttt{2}).
} 
\label{tab:all-joint-val}
\end{table*}

%% file: tables/spatial-anno-template-wise.tex
\begin{table*}[]
\centering
\resizebox{0.8\textwidth}{!}{%
\begin{tabular}{@{}llccc@{}}
\toprule
Category                 & Template                                       & \begin{tabular}[c]{@{}c@{}}Num videos\\ selected\end{tabular} & \begin{tabular}[c]{@{}c@{}}Num frames\\ annotated\end{tabular} & \begin{tabular}[c]{@{}c@{}}Total duration\\ annotated (in secs)\end{tabular} \\ \midrule
\multirow{9}{*}{Objects} & Where is object X before / after event Y?      & 10                                                            & 127                                                            & 25.4                                                                         \\
                         & Where is object X?                             & 10                                                            & 121                                                            & 24.2                                                                         \\
                         & What did I put in X?                           & 10                                                            & 155                                                            & 31                                                                           \\
                         & How many X's? (quantity)                       & 10                                                            & 157                                                            & 31.4                                                                         \\
                         & What X did I Y?                                & 10                                                            & 119                                                            & 23.8                                                                         \\
                         & In what location did I see object X?           & 10                                                            & 140                                                            & 28                                                                           \\
                         & What X is Y?                                   & 10                                                            & 152                                                            & 30.4                                                                         \\
                         & State of an object                             & 10                                                            & 149                                                            & 29.8                                                                         \\
                         & Where is my object X?                          & 10                                                            & 149                                                            & 29.8                                                                         \\ \midrule
Place                    & Where did I put X?                             & 10                                                            & 96                                                             & 19.2                                                                         \\ \midrule
\multirow{3}{*}{People}  & Who did I interact with when I did activity X? & 10                                                            & 100                                                            & 20                                                                           \\
                         & Who did I talk to in location X?               & 10                                                            & 142                                                            & 28.4                                                                         \\
                         & When did I interact with person with role X?   & 10                                                            & 150                                                            & 30                                                                           \\ \midrule
\multicolumn{2}{c}{(mean $\pm$ std)}                                                 & (10 $\pm$ 0)                                                           & (135.15 $\pm$ 19.85)                                                          & (27.03 $\pm$ 3.97) \\ \midrule       
\multicolumn{2}{c}{Total}                                                 & 130                                                           & 1757                                                           & 351.4                                                                        \\ \bottomrule
\end{tabular}%
}
\caption{\textbf{Selection of videos for spatial-annotation for NLQ task.} We uniformly sample $10$ videos across $13$ question templates and annotate spatial bounding-boxes for each ground-truth extent. In total, we annotated bounding-boxes for $1757$ frames or $351.4$ secs of video footage where the answers to NLQ queries can be found.}
\label{tab:spatial-anno-template-wise}
\end{table*}

%% file: tables/nlq-spatial-zero-shot-full.tex
\begin{table*}[]
\centering
\resizebox{0.7\textwidth}{!}{%
\begin{tabular}{@{}c|cccc|ccc@{}}
\toprule
\multicolumn{1}{l|}{}           & \multicolumn{4}{c|}{Spatio-temporal}                                                                                          & \multicolumn{3}{c}{Temporal}                                                                                                                                             \\ \midrule
\multirow{2}{*}{Model}  & \multirow{2}{*}{\begin{tabular}[c]{@{}c@{}}spatial\\ branch\end{tabular}} & \multicolumn{2}{c}{stIoU=0.3} & \multirow{2}{*}{\begin{tabular}[c]{@{}c@{}}mean\\ stIoU\end{tabular}}      & \multicolumn{2}{c}{tIoU=0.3} & \multirow{2}{*}{\begin{tabular}[c]{@{}c@{}}mean\\ tIoU\end{tabular}}   \\ 
\cmidrule(lr){3-4} \cmidrule(lr){6-7} `
&       & R@1      & R@5        &     & R@1 &R@5 &                         \\ \midrule
NLQ-only & N/A & - & - & - & 9.30 & 18.60 & 5.35 \\ \midrule
\multirow{4}{*}{\begin{tabular}[c]{@{}c@{}}MINOTAUR\\ (All-Tasks)\end{tabular}}& random boxes  & 0 $\pm$ 0         & 0 $\pm$ 0             & 0.40 $\pm$ 0.04 &\multirow{4}{*}{\textbf{11.63}} &\multirow{4}{*}{\textbf{20.93}} & \multirow{4}{*}{\textbf{8.35}} \\ 
& \begin{tabular}[c]{@{}c@{}}random centered\\ boxes\end{tabular}           & 0 $\pm$ 0         & 0.47 $\pm$ 0.38       & 1.25 $\pm$ 0.03 & & & \\
\cmidrule(l){2-5}
 & All-Tasks & \textbf{2.33}        & \textbf{4.65}            & \textbf{2.27} & & &  \\
\bottomrule
\end{tabular}%
}
\caption{\textbf{Zero-shot spatio-temporal grounding on NLQ.} In addition to the Table \textcolor{red}{3} in the main paper, we report recall numbers on temporal metrics. We evaluate spatio-temporal predictions on NLQ by annotating a subset of validation videos ($=130$), and comparing it with random baselines. We observe non-trivial performance despite the model not trained on (language query, spatio-temporal target).}
\label{tab:nlq-zero-shot-full}
\vspace{-8pt}
\end{table*}

%% file: tables/vos.tex
\begin{table}[]
\centering
\begin{tabular}{@{}l|c|ccc@{}}
\toprule
Method                    & Pretraining                                                     & $\mathcal{J}$\&$\mathcal{F}$  & $\mathcal{J}$     & $\mathcal{F}$     \\ \midrule
\multirow{2}{*}{STM}      & None                                                            & 62.8  & 60.6  & 64.9  \\ \cmidrule(l){2-5} 
                          & COCO                                                            & 75.8  & 73.6  & 78.0  \\ \midrule
\multirow{5}{*}{MINOTAUR} & ImageNet                                                        & 16.34 & 12.11 & 20.58 \\ \cmidrule(l){2-5} 
                          & MDETR                                                           & 70.95 & 69.05 & 72.86 \\ \cmidrule(l){2-5} 
                          & \begin{tabular}[c]{@{}c@{}}Ego4D\\ (\textit{All-Tasks})\end{tabular} & 73.15 & 71.32 & 74.98 \\ \bottomrule
\end{tabular}%
\caption{\textbf{Semi-Supervised Video Object Segmentation results on VISOR \cite{VISOR2022} validation set.} We compare with the STM \cite{oh2019video} baseline reported in VISOR \cite{VISOR2022}. We observe that, 1) 
our method (\textit{All-Tasks}) performs comparably to STM \cite{oh2019video} which is pre-trained on additional segmentation data (from COCO), and 2) our Ego4D (\textit{All-Tasks}) initialization outperforms MDETR \cite{kamath2021mdetr} initialization, which shows the generic and expressive nature of our features for other tasks at least in egocentric domain.
}
\label{tab:vos}
\end{table}

%% file: tables/ablation-sampling-strategy.tex
\begin{table*}[t]
\centering
\begin{tabular}{@{}l|cccc|cccc|cccc@{}}
\toprule
\multirow{3}{*}{Strategy} & \multicolumn{4}{c|}{VQ2D} & \multicolumn{4}{c|}{NLQ} & \multicolumn{4}{c}{MQ} \\ \cmidrule(l){2-13} 
 & \multirow{2}{*}{tAP25} & \multirow{2}{*}{stAP25} & \multirow{2}{*}{rec} & \multirow{2}{*}{Succ} & \multicolumn{2}{c}{tIoU=0.3} & \multicolumn{2}{c|}{tIoU=0.5} & \multicolumn{2}{c}{tIoU=0.3} & \multicolumn{2}{c}{tIoU=0.5} \\ \cmidrule(l){6-13} 
 &  &  &  &  & R@1 & R@5 & R@1 & R@5 & R@1x & R@5x & R@1x & R@5x \\ \midrule
Concat & 0.41 & \textbf{0.20 }& \textbf{27.1} & 59.3 & 6.20 & 17.86 & 3.98 & 10.92 & 31.54 & 53.51 & 21.97 & 39.64 \\
Round-Robin & \textbf{0.41} & 0.19 & 26.4 & \textbf{60.2} & \textbf{7.56} & \textbf{20.06} & \textbf{4.62} & \textbf{11.80} & \textbf{33.28} & \textbf{54.93} & \textbf{24.09} & \textbf{40.27} \\ \bottomrule
\end{tabular}%
\caption{\textbf{Sampling strategy for multi-task learning (Round Robin v/s Concat).} We train our All-Tasks (AT) approach using different sampling strategies and evaluate on validation set. We found that \textit{Round-Robin} performs better than \textit{Concat} on $10$ out of $12$ metrics across the three tasks.}
\label{tab:ablation-sampling-strategy}
\end{table*}

%% file: tables/ablation-fps-5-mq.tex
\begin{table}[H]
\centering
\resizebox{\columnwidth}{!}{%
\begin{tabular}{@{}l|c|cccc@{}}
\toprule
 & \multirow{3}{*}{Methods} & \multicolumn{4}{c}{MQ} \\ \cmidrule(l){3-6} 
 &  & \multicolumn{2}{c}{tIoU=0.3} & \multicolumn{2}{c}{tIoU=0.5} \\ \cmidrule(l){3-6} 
 &  & R@1x & R@5x & R@1x & R@5x \\ \midrule
\multirow{2}{*}{Single-Task} & fps 5 & 17.55 & 40.82 & 12.12 & 28.49 \\
 & fps 1 & \textbf{33.64} & \textbf{55.38} & \textbf{23.86} & \textbf{39.48} \\ \bottomrule
\end{tabular}%
}
\caption{\textbf{MQ: fps 5 v/s fps 1.} We obtain better performance on MQ task with lower fps, since lower fps allows capturing longer temporal context, and this works well for MQ since out of the three tasks, MQ has the longest ground-truth extents.}
\label{tab:ablation-fps-5-mq}
\end{table}

%% file: tables/ablation-step-size-mq.tex
\begin{table}[H]
\centering
\resizebox{\columnwidth}{!}{%
\begin{tabular}{@{}l|c|cccc@{}}
\toprule
 & \multirow{3}{*}{\begin{tabular}[c]{@{}c@{}}Step size\\ $k_{step}$\end{tabular}} & \multicolumn{4}{c}{MQ} \\ \cmidrule(l){3-6} 
 &  & \multicolumn{2}{c}{tIoU=0.3} & \multicolumn{2}{c}{tIoU=0.5} \\ \cmidrule(l){3-6} 
 &  & R@1x & R@5x & R@1x & R@5x \\ \midrule
\multirow{3}{*}{All-Tasks} & 200 & 30.49 & 51.93 & 20.01 & 37.31 \\
 & 100 & 33.28 & 54.93 & 24.09 & 40.27 \\
 & 50 & \textbf{34.82} & \textbf{56.68} & \textbf{25.58} & \textbf{42.49} \\ \bottomrule
\end{tabular}%
}
\caption{\textbf{Effect of step size $k_{step}$ during evaluation on MQ.} We observe that with more overlap between sliding windows (or lower step size $k_{step}$), we obtain better performance.}
\label{tab:ablation-step-size-mq}
\end{table}